Evaluating GenAI for Simplifying Texts for Education: Improving Accuracy and Consistency for Enhanced Readability

Stephanie L. Day[1]
Jacapo Cirica[1]
Steven R. Clapp[1]
Veronika Penkova[1]
Amy E. Giroux[1]
Abbey Banta[1]
Catherine Bordeau[1]
Poojitha Mutteneni[1]
Ben D. Sawyer[1]

[1]University of Central Florida

Corresponding Author: Stephanie Day; stephanie.day@ucf.edu, 4000 Central Florida Blvd, P.O. BOX 162993, Orlando, FL 32816-2993

This work was supported by the Readability Consortium and with gift funds from Adobe, Inc. All authors approve this manuscript in its current form and no conflicts of interest exist. This study did not include human subjects.

**Abstract**

**Background:** Generative artificial intelligence (GenAI) holds great promise as a tool to support personalized learning. Teachers need tools to efficiently and effectively enhance content readability of educational texts so that they are matched to individual students' reading levels, while retaining key details. Large Language Models (LLMs) show potential to fill this need, but previous research notes multiple shortcomings in current approaches.

**Objective:** In this study, we introduced a generalized approach and metrics for the systematic evaluation of the accuracy and consistency in which LLMs, prompting techniques, and a novel multi-agent architecture to simplify sixty informational reading passages, reducing each from the twelfth grade level down to the eighth, sixth, and fourth grade levels.

**Method:** We calculated the degree to which each LLM and prompting technique accurately achieved the targeted grade level for each passage, percentage change in word count, and consistency in maintaining keywords and key phrases (semantic similarity).

**Results and Conclusions:** One-sample t-tests and multiple regression models revealed significant differences in the best performing LLM and prompt technique for each of the four metrics. Both LLMs and prompting techniques demonstrated variable utility in grade level accuracy and consistency of keywords and key phrases when attempting to level content down to the fourth grade reading level. These results demonstrate the promise of the application of LLMs for efficient and precise automated text simplification, the shortcomings of current models and prompting methods in attaining an ideal balance across various evaluation criteria, and a generalizable method to evaluate future systems.

Key Words: artificial intelligence, education, elementary education, high school, computer supported collaborative learning

## 1. Introduction

Reading comprehension and fluency are critical skills required for both academic success and life beyond. Post-COVID-19 reading scores in some grade levels have displayed the largest drops since 1990 across low-, middle-, and high-performing students, with low-performing students showing the sharpest decreases (Irwin et al. 2023). In typical general education classrooms, reading skills may vary greatly across classrooms and grade levels (Kärner et al., 2021). While the highest achieving students may read well above their current grade level, others may struggle with the readability of grade-level-appropriate content (Firmender et al., 2013). These gaps present a challenge for teachers to design and deliver instruction and materials that effectively meet the diverse needs of all students to support stronger reading comprehension and learning outcomes. Furthermore, the achievement gap may widen between students based on race, ethnicity, and socioeconomic status (Reardon, 2018). The National Center for Education Statistics (2022) reported approximately 10.6% of students (5.3 million) in the U.S. in 2021 were English learners (ELs) and there are significant disparities in achievement outcomes (Calderón et al., 2011). Research on the use of differentiated instruction (DI) has consistently shown that personalized instruction and materials are correlated with improved academic outcomes, confidence, school well-being, and engagement (Connor et al. 2013; Smale-Jacobse et al. 2019; Pane et al., 2017; Pozas et al. 2021). However, creating instructional materials that support readability across multiple grade levels is complex and well exceeds the time most educators have available.

Generative AI (GenAI) for the classrooms holds great promise as a tool for creating personalized learning experiences that could benefit both teachers and students with enhanced readability to support better fluency and comprehension. However, many concerns exist surrounding the quality and accuracy of GenAI content (Zastudil, C., Rogalska, M., Kapp, C., Vaughn, J., & MacNeil, S., 2023). Textual GenAI can generate text based upon user-inputted prompts, often leveraging large language models (LLMs). These advanced neural networks are trained on vast amounts of text data to generate written content by predicting the next word in a sequence, enabling them to perform various natural language processing (NLP) tasks. Notably, GenAI systems architecture is often a mix of neural networks and more traditional deterministic systems. While emerging research has demonstrated the potential utility of GenAI to simplify educational texts, many challenges and questions remain on the best approaches and metrics in which to evaluate the accuracy and effectiveness of GenAI in this context (Farajidizaji et. al. 2024; Huang et al., 2024a). In this exploratory study, we evaluated the use of three LLMs, four prompting techniques, and a new multi-agent architecture to relevel texts (text simplification) on various informational topics across grade levels to assess consistency in maintaining key ideas, vocabulary, grade level, and passage length (word count).

## 1.1 Personalized Learning

Students reading both above and below grade level often become frustrated and disengaged if content is not readable, either too challenging or not challenging enough (Little et al., 2014). Vygotsky's (1978) Zone of Proximal Development (ZPD) offers a theoretical framework to guide text leveling decisions within a differentiated approach. ZPD refers to the space between what a learner can do independently (actual developmental level) and what they can only do with guidance and support (tasks beyond current capabilities). When students receive instruction and materials within their ZPD, they are challenged just enough to acquire new reading skills and strategies without becoming overwhelmed or disengaged. Providing materials that are at, or just above an individual's reading level may be particularly important for developing and struggling readers as well as second language learners. Research has shown that students provided with leveled texts, which were mildly to moderately challenging, displayed improvements in reading strategies, phonetic decoding, word recognition, and fluency (Alowais, 2021; Denton et al., 2014). Studies show that increased text difficulty can negatively affect comprehension because cognitive resources are focused on decoding individual words rather than connecting ideas (Amendum, Conradi, & Hiebert, 2017). The use of leveled texts not only builds confidence and motivation but also facilitates a gradual progression to more complex texts as students' reading abilities improve over time (Torgesen et al., 2017). Thus, developing and providing readable curricula at various reading levels can be beneficial for both teachers and students, yet a general practice in schools is to use curricula that does not offer alternative versions across multiple grade levels of difficulty (Alowais, 2021), which limits teachers' ability to provide differentiated instruction (DI).

## 1.2 Text Simplification

Text simplification, also referred to as text leveling (or releveling), is complex. It can take significant time and training for a human to understand and is labor-intensive even for experts. Manual text leveling is a technique in which a human manually adjusts the complexity of a text to match different students' reading abilities by simplifying more sophisticated grammatical constructions, shortening sentences, and reducing vocabulary load, while also aiming to maintain key ideas. The greater the leap – say, simplifying a text from 12th grade to 4th grade

— the more time may be required. Some topics may also take longer to relevel, such as more technical scientific topics. Texts that include more prevalent specialized terminology, multisyllabic vocabulary, and intricate syntactic structures are more challenging to relevel and require more time to achieve downgraded readability levels without significantly impacting the meaning of the text.

Automatic text releveling through simplification is a well-known and non-trivial task in natural language processing (NLP), a field focused on the interaction between computers and human language, where the original text is adapted to make it easier to understand by reducing linguistic complexity. Linguistic complexity encompasses various attributes that can make a text more challenging to understand, including sentence length, vocabulary difficulty, syntactic structure, morphological variations, discourse organization, and semantic density (Siddharthan, 2014). When linguistic complexity is reduced, the desired output is easier to understand and facilitates readability for a wide range of audiences (Alonzo et al., 2020; Evans et al., 2014; Rello et al., 2013; Watanabe et al., 2009).

Teachers are not usually trained in text leveling and related adaptation strategies, thus text simplification does not maximize the productive use of teacher skill or time (Crossley et al, 2011; Jin & Lu, 2017). Nonetheless, this is a need that most teachers face within their classrooms. Programs like CohMetrix (McNamara & Graesser, 2012) and the Lexile Analyzer (Lennon & Burdick, 2004) (most commonly used by educators) provide text difficulty metrics, but do not offer the ability to relevel a text.

## 1.3 Generative AI

GenAI tools offer an attractive and potentially efficient solution to enhance readability: the ability to quickly assess text difficulty and regenerate equivalent content at various levels, enabling teachers to provide each student with suitably challenging materials without undue labor and costs (Pratama et al., 2023; Chen et al., 2020). While traditional software tools are mainly focused on analysis, prediction, and decision-making tasks, GenAI can generate new content conforming to patterns in real existing data (García-Peñalvo et al., 2023). This advancement has been facilitated by extensive training datasets, enhanced computational power, and the advent of transformer architectures (Vaswani et al., 2017). Models trained on large datasets of existing content, including generative adversarial networks (GANs) or variational autoencoders (VAEs), can also be trained to respond to general-language conversational 'chat'-like instructions. The result is that prompts created by non-experts can result in contextually-appropriate and coherent written content (Naveed, 2023). The large-scale generative foundation models, such as the widely recognized ChatGPT (OpenAI, 2023), have recently very rapidly advanced the state-of-the-art in performance across various natural language tasks, including text simplification (Feng et al., 2023). Further, as trust in GenAI technology for classroom use is lacking, developing personalized learning technology must be trustworthy (Lee & See, 2004), simultaneously satisfying multiple goals including simplification of some content, and contextual preservation of main ideas. Technology that is not trustworthy in this way will not achieve technology acceptance (Sawyer, Miller, Canham & Karwowski, 2021).

The goal of personalized learning technology is to improve the efficiency of learning (Li & Wong, 2021), but commercially available tools currently remain insufficient when it comes to text simplification. Simply downgrading a text to reduce difficulty is only part of the challenge within a classroom. While a teacher may want to offer a text at various levels, each version must be consistent in key ideas or even specific vocabulary. For example, all students will need to complete the same version of an end-of-unit test, irrespective of their reading level, thus while the reading level of the material can vary, certain vocabulary and concepts would need to be present across all versions. A critical consideration in the present work will be to evaluate what information may be lost in the process of simplifying the text, and to develop tools that can maintain a desired level of consistency. Thus, two primary questions emerge in this work: 1. Can LLMs accurately relevel texts to the targeted grade level and 2. Can LLMs maintain consistency in key ideas and vocabulary when simplifying a text?

### 1.4 Related Research

Patel et al. (2022) pioneered the study of LLM-based textual simplification for educational purposes, focusing on mathematical problems using GPT-3. Their work explored the LLM's text prompts and the few-shot learning method with promising results, such as improvements in readability metrics. However, there are still challenges, such as the model sometimes giving irrelevant results (noise), requiring manual processing to correct errors and improve quality, inconsistencies with information presented, struggling with complex math symbols, and inconsistencies across grade levels or curricula. Moreover, the task is restricted to the simplification of mathematical problems and uses an outdated version of the LLM.

An initial investigation by Farajidizaji et. al. (2024) showed that text readability could be modified to a targeted grade level using zero-shot large language models. Zero-shot learning refers to the model's ability to perform a task without having seen any examples during training, as opposed to few-shot learning, where the model is given a small number of examples to learn from. The results showed that both ChatGPT and Llama-2 can relatively control readability levels (defined by Flesch Reading Ease scores). Further, ChatGPT performed better through a two-step process which included generating paraphrases to improve accuracy. While this study demonstrated the ability of LLM's to guide text difficulty towards desired levels, exact precision was a challenge. Further, it was also found that semantic similarity between the source text and modified text decreased as the grade-level range was increased, indicating poor consistency.

Huang and colleagues (2024a) introduced a leveled-text generation task. Their goal was to rewrite different educational passages to targeted grade levels while retaining the main meaning of the content using different LLMs. The study was performed with GPT-3.5, LLaMA-2 70B, and Mixtral 8x7B, and they used zero-shot and few-shot prompting. Their findings showed that few-shot prompting significantly enhanced grade-leveled manipulation and information retention. LLaMA-2 70B excelled in adapting text difficulty, while GPT-3.5 was superior at maintaining meaning. However, they also found that paraphrasing and inconsistent edits from the original texts persisted. The models tended to shorten the leveled text from the original 825 words to as low as 350 words. Text simplification is different from paraphrasing and sentence compression, and it requires the ability to retain the most important information, including keywords and phrases (Barzilay & Elhadad, 2003; Knight & Marcu, 2002). Further disadvantages included inaccurate information and loss of key information. These results suggest that a more intuitive model may be needed that could identify the important details of the text to maintain consistency.

Overall, these studies show the potential application of prompt engineering techniques to LLMs to adjust the grade level of a source text, however, significant challenges remain. First, while the models can push the grade level difficulty in the desired direction, they often fail to precisely match the targeted grade level (Huang et al., 2024a). Secondly, there are often changes to the word count, which creates potentially unbalanced versions of a text and results in paraphrasing, thus losing potentially important information. Semantic changes introduced by GenAI, such as paraphrasing or replacing important keywords in the source text, can result in a loss of content. Maintaining specific vocabulary words, concepts, and details is of particular importance, as significant changes to content reduce the overall quality of the content, impacting readability and learning. Findings also suggest a greater loss in semantic similarity as the range between the source text and the desired level of the modified text increases.

Determining the extent to which key information was lost or changed during the simplification process and finding effective prompting techniques to better maintain consistency is therefore a primary focus of this present work. However, all existing proposed solutions and experiments have relied solely on prompt engineering to solve the task, ignoring more advanced techniques such as Chain-of-Thought (Wei et al.,2022) or Prompt Chaining (Wu, Terry & Cai, 2022). In the present work we will specifically engage with these more elaborate architectures and frameworks, which have been neglected to date in the literature.

**1.5 Research Aims**

In general, the application of GenAI for educational purposes such as text simplification is in its infancy, and ongoing innovation and investigation is needed to improve grade level accuracy, consistency in maintaining key information, and overall passage structure. Therefore, in this exploratory study, we first investigated the application of unexplored architectures and prompt techniques to Large Language Models (LLMs). We then assessed differences in their ability to match targeted grade levels, maintain consistency of key vocabulary and key concepts (phrases), and word count.

As a first step in this work, we proposed an enhanced approach to generate informational passages across various topics and grade levels. In contrast to the task performed by Huang and colleagues (2024a), we not only aimed to relevel 12th grade-leveled passages to a specific lower grade level, we also aimed to maintain important keywords and details, and to retain word count/length for improved readability. While maintaining word count may not necessarily always be of importance, constraining word count may be an effective method in which to prevent LLMs from paraphrasing a text when downgrading the grade level.

In the current experiment, we evaluated the performance of three Large Language Models (LLMs), namely GPT-4-Turbo (OpenAI, 2023), Claude 3 (Anthropic, 2024), and Mixtral 8x22B (Jiang et al., 2024), to generate lower grade-leveled content through three recognized prompt engineering techniques: Prompt Chaining, Chain-of-Thought, and Directional Stimulus Prompting. As the baseline of the study, we tested the releveling task first with a zero-shot prompt. We also introduced and evaluated a new framework approach using an agentic-based architecture. We assessed differences in performance between these LLMs and prompting techniques across a set of text evaluation metrics that measured how closely downgraded versions of a 12th grade source text matched the targeted grade level, accuracy of maintaining key words, semantic similarity of important details, and the percent change in word count. The following questions guided this experiment:

1. Which LLM and prompting technique(s) had the best accuracy at downgrading passages to the targeted grade level?

2. Which LLM and prompting technique(s) had the highest consistency in maintaining keywords?

3. Which LLM and prompting technique(s) demonstrated the most consistency in maintaining semantic similarity across downgraded passages?
   a. Does semantic similarity decrease as a function of the grade level difference between the original text and the simplified text?

4. Which LLM and prompting technique(s) performed best at maintaining word count?

## 2. Methods

The primary aim of this work was to relevel a set of 12th grade human-generated passages down to the 8th grade, 6th grade, and 4th grade levels through the use of LLMs and prompting techniques to evaluate the accuracy of the releveling and consistency in content and structure. We initially used several advanced prompting techniques not explored in previous work (Huang et al., 2024a); however, as informed by prior work suggesting shortcomings, we also introduced a multi-agent framework. We assessed whether passages were accurately releveled at the targeted grade level, maintained the passage structure and word count, and importantly, retained keywords and key phrases in relation to the 12th grade baseline passage on each topic. All these techniques aim to improve the reasoning capabilities (Lewkowycz et al., 2022; Kojima et al., 2022) of LLMs and are then integrated into our multi-agent framework.

**2.1 12th Grade Passage Dataset:**

First, we developed sixty 300-word passages (+/- 10%) to serve as a baseline in which each LLM would be instructed to simplify. All passages were created at a 12th-grade reading level as measured by the Flesch-Kincaid Grade Level (FKGL) scale. Five of the passages were biographies, nine humanities passages (how we understand the world), ten passages about current events (within the last 50 years), fourteen science/psychology passages, seven passages about U.S. history, and fifteen passages about world history. We provide a sample of these passages on each topic in the Supplemental Materials Section 1. The passages were written by two experienced educators and curriculum developers. The full dataset of passages is available in our repository (Day et al., 2025). We included a variety of content and subject areas to create a more robust experiment in which to provide more generalizable results.

**2.1.1 *Keywords and Key Phrases.*** As stated above, a primary aim of this work was not simply to relevel a text on a particular topic, but to retain content - keywords, and key phrases, across grade levels. Thus, our content writers identified keywords and phrases in each of the 12th grade-leveled baseline passages. Keywords were specific vocabulary words that we wanted to maintain across the text simplification process. For example, in a passage about ancient Egypt, a teacher might wish to retain the word "hieroglyphic" rather than allowing the LLM to attempt to simplify the word to "symbol" as students will be tested on the meaning of hieroglyphic. Keywords were *not* to be modified or replaced in the simplification process. While in some cases keywords were identified as a single word, keywords could also be a group of 2 to 5 words such as proper nouns (e.g., Mount Vesuvius). The number of keywords retained per passage ranged from 8 to 23 (M = 15.06).

Unlike keywords, key phrases represented important concepts such that simplification of the exact text would be acceptable as long as the main idea of the key phrase was maintained. Key phrases included key concepts that help the reader understand main ideas, series of events, real-word relevance, and cause and effect. Phrases were typically several words or a full sentence, and the number of retained key phrases per passage ranged from 9 to 19 (M = 13.22). Key phrases could be altered to suit decreasingly skilled audiences through sentence structure simplification (shortening or breaking up into multiple shorter sentences) and vocabulary modification. Sample keywords and key phases for one passage can be seen in Table 1.

In some cases, we also asked the LLMs themselves to select them. LLM selection occurred in three cases: (1) **Prompt Chaining**- as certain keywords and key phrases are generated for the second prompt, (2) **CoT**- as one of the steps is to select certain key words and key phrases, and (3) **Multi-Agent** workflow- as an agent is the *Selector* assigned to select the concepts and words not to be changed (max. 5 each). For evaluation purposes, however, we only took into account the accuracy of the models to maintain the keywords and key phrases selected by the human experts.

**2.1.2. *Passage Word Count***. For this task, the original passages were set at a desired word count. All passages were approximately 300 words (+/-10%, range 270-330) divided into 4 paragraphs, each consisting of approximately 75 words (+/- 5%, range 71-79 words) to support readability. We chose 300 words as a starting point for this work to assess quality and consistency in shorter reading passages, which are passages that might appear on a reading comprehension test, for example. While evaluating the consistency, content retention, and quality of longer texts is a later goal of this work, 300-word passages are long enough to provide sufficient details on various topics in which to draw comparisons across grade levels and short enough to allow for efficient comparison and evaluation. Starting with a shorter length will serve as a good baseline in which to build upon to assess consistency and quality issues as word count is increased in future work. An important requirement of the LLM to level the passage was to keep the word count as close to the original without changing the content to avoid paraphrasing.

**2.2 Large Language Models (LLMs):**

We evaluated the capability of three LLMs via API: GPT-4-Turbo, Claude 3 (Anthropic, 2024) and Mixtral 8x22B. Specifically, we used *gpt-4-turbo* (OpenAI, n.d.)*, claude-3-opus-20240229* (Anthropic, n.d.) and *Mixtral-8x22B-v0.1* (Mistral AI, n.d.). It may be important to note that code for GPT-4-Turbo and Claude3 are not open sourced while Mixtral 8x22B has weights that are open-sourced (HuggingFace, n.d.), allowing future solutions to further finetune the zero-shot solution specifically for text simplification.

These LLMs were chosen because at the time of the present experiment they appeared to be the models that could achieve the most competitive performance. In fact, these model families had the best scores on AlpacaEval 2.0 (Dubois et al., 2024) and MTBench (Zheng et al., 2023) which are two leading benchmarks for assessing the alignment of LLMs with human preferences.

**2.3 Known Prompting Techniques**

As observed in previous work, LLMs require natural language prompts to control the generated paraphrases (Farajidizaji et al., 2024), and we therefore used the following prompting strategies for the first part of the experiments. We started, however, from a zero-shot solution as a baseline for the subsequent experiments (see Supplemental Materials Section 3 Prompts for the full prompts used).

**2.3.1 Zero-shot prompting:** Zero-shot prompting refers to using a prompt that does not include any examples or demonstrations when interacting with the model. In this approach, the prompt directly instructs the model to perform a task without providing any examples to guide it. Prompts corresponding to each target grade level can be found in Supplemental Materials Section 3. The prompts are selected in relation to the LLM of use.

**2.3.2 Directional Stimulus Prompting** (Li et al., 2023). We also assessed grade level accuracy and content retention using Directional Stimulus prompting (DSP). We started with this technique as we were primarily interested in evaluating the behavior of LLMs with explicit human-annotated key words and key phrases. DSP relies on "hints" from the transformer with details; e.g., in our case the key words and phrases. This helps the Model understand what the important sections that need to be maintained are. In this research we followed two approaches; the first one passed the passage through another prompt that is responsible for highlighting keywords and phrases. Then, the selected keywords and phrases were input into our next prompt, which maintained the keywords while reducing the grade level of the generated passage (Figure 1). The prompt used for the DSP can be found in Supplemental Materials Section 3.

**2.3.3 Chain-of-Thought (CoT) Prompting**: Chain-of-thought (CoT) prompting, introduced by Wei et al. (2022), enhances complex reasoning abilities by incorporating intermediate reasoning steps. The aim of CoT was to preserve the essential content of the text and reduce the noise of the regenerated text. "Noise" refers to irrelevant, incorrect, or extraneous information generated by the LLM. We fed the input with some examples from the dataset, extrapolated through the Automatic CoT prompting technique. Auto-CoT automates the process of instructing LLMs using the "Let's think step-by-step" prompt to generate reasoning chains, eliminating the need for manually crafted examples in chain-of-thought prompting. This approach, introduced by Zhang et al. (2022), addresses the potential errors in individual reasoning chains by increasing the diversity of demonstrations. The technique is not so dissimilar to Prompt Chaining; however, the difference is that we combined it with few-shot prompting and also included the selection of the most relevant keywords and phrases as an intermediate step in order to avoid the conceptual loss or hallucinations observed in previous attempts (Figure 2).

**2.3.4 Prompt Chaining** (Wu, Terry and Cai, 2022) was first used to improve the model's understanding of context by providing it with a series of prompts or steps guiding the LLM to perform specific text transformations. This methodology is used to break down complex tasks into smaller, manageable sub-tasks. A prompt chain typically consists of several prompts, either specific or general-purpose, each designed to serve a single function. The output of one prompt becomes the input for the next. In our task, the first prompt included a series of guided stages such as identifying longer sentences and words. Then, it is followed by a second prompt whose task is breaking down the previous sentences and words into shorter, simpler sentences while retaining the main meaning (Figure 3).

**2.3.5 Multi-Agent Architecture:** In addition to evaluating previously tested prompting techniques, we also propose our conversational problem-solving framework, leveraging the AutoGen *dynamic group chat* communication pattern, which has previously been used to solve mathematical problems (Wu et al., 2023a). We chose to test this approach due to limitations observed in previous work (Farajidizaji et al., 2024; Huang et al., 2024b; Patel et al., 2022).

We adapted this framework to modify the grade level of our informational baseline passages by involving several sub-agents. We used the chat model *gpt-4-1106-preview* for running each agent. We also experimented with using *gpt-3.5-turbo-instruct*. It is important to note that this solution relies on the AutoGen framework, which currently only supports OpenAI models (GPT) at the time of this experiment. We used a vertical structure (Masterman et al., 2024), where one agent acts as a leader and the other agents report directly to it. The conversation is shared between all agents and there is a clear division of labor between the collaborating agents. We created a modular framework, employing five separate models. First, a *Manager* agent, which is an instance of the *GroupChatManager* class, performs the following three operations: (1) select a single speaker (in this case *Selector*) and broadcast the selected speaker's message to all other agents, (2) the *Selector* reads the source text and develops a strategy to select the most relevant keywords and phrases, and (3) Information is then passed to an agent *Writer,* which is responsible for rewriting the text to the specified grade level and to maintain formatting. A *Calculator,* which is denoted as an executor agent, is a customized UserProxyAgent responsible for executing Python code to calculate the FKGL score, and the word count generated by the Writer. Finally, the output is provided to the Editor agent which evaluates the text and suggests possible changes.

Designing a multi-agent workflow for a specific task requires several decisions, such as determining the number of agents to include, assigning roles and capabilities to each agent, defining their interactions, and deciding which parts of the workflow should be automated. For our case and specific task, we propose a straightforward four-agent system, illustrated in Figure 4 and the workflow is detailed in the Supplemental Materials Section 2.

## 2.4 Grade-leveled Passage Generation (Text Simplification):

To generate passages at lower grade levels for this experiment, we defined our enhanced grade-leveled passage generation task (Figure 5) as:

"*Given a source text with 4 subparagraphs* $T_i$ $(y_{i,1}, y_{i,2}, y_{i,3}, y_{i,4})$, *a function* $F$ *for calculating readability score, a desired word length* $L(\lambda_1, \lambda_2, \lambda_3, \lambda_4)$, *and* $R$ *pre-specified readability scores* $\{r_1, r_2, r_3, ...., r_n\}$, *generate* $R_i \{x_1, x_2, x_3, .. x_n\}$ *versions of the source text* $T_i$ *such that it has* $F(x_i) = r_i$ *and length* $\approx L$ *words.*"

In this work, we set $R$ = 3 with $r_1$ = 8, $r_2$ = 6, $r_3$ = 4. $L = 300 \, words$ ($\pm$ 10%) and each subparagraph ($\lambda_1, \lambda_2, \lambda_3, \lambda_4$) is equal to approximately 75 words with a tolerance of $\pm$ 10%.

The **Flesch-Kincaid Grade Level** (FKGL; Flesch,1948) was utilized as our metric of grade level difficulty (readability function $F$ above). While we acknowledge there are additional metrics of text cohesion and coherence for assessing the readability of a text, in the present study, we focus on FKGL as it is one of the most common and easily interpretable metrics of grade level readability in educational research and is open-sourced. FKGL considers the total number of words, sentences, and syllables to calculate the recommended U.S. grade level of a text. It is read from the lowest to highest value. For example, FKGL=8 reads as if the content can be understood by a student in 8th grade. In this study, we generated content at the 12th, 8th, 6th, and 4th-grade levels according to FKGL (Lipovetsky, 2023). The target readability scores are selected as the midpoint values for each FKGL range. It is measured with the formula:

$$\textbf{FKGL} = 0.39 \left(\frac{n_w}{n_{se}}\right) + 11.8 \left(\frac{n_{sy}}{n_w}\right) - 15.59$$

Where $n_w$ indicates the total number of words, $n_{se}$ indicates the total number of sentences, and $n_{sy}$ indicates the total number of syllables. FKGL is chosen as a straightforward readability metric due to its easily interpretable score ranges and its strong correlation with human comprehension as assessed by reading tests (DuBay, 2007).

## 2.5 Releveled Passage Evaluation Metrics

As informed by challenges in prior work, we chose to focus on four primary evaluation metrics in which to assess LLM and prompt accuracy and consistency: 1) Grade Level Accuracy: The degree to which each simplified passage measured at the targeted FKGL, 2) Keyword Accuracy The degree to which each simplified passage consistently maintained keywords, 3) Semantic Similarity: Consistency of key phrases in each simplified passage, and 4) Word Count Change %: The change in word count relative to the original number of words in a passage.

**2.5.1 Grade-Level Accuracy:** A primary aim of this work was to simplify informational passages from 12th grade down to the 8th, 6th, and 4th grade level. To assess grade level accuracy, within each LLM and prompting technique combination, we calculated the FKGL of each releveled passage. Within each grade level, FKGL scores range from _.00 to _.99, representing the lower and upper bounds of difficulty within each grade level. Thus, for the purpose of the current experiment, we opted to compare the FKGL of each simplified passage to the midpoint within each targeted grade level. For example, the midpoint FKGL between 6.00-6.99 is 6.5. One-sample t-tests were performed on the FKGL scores derived from each LLM/prompt combination's set of sixty simplified passages. We then examined which averages were statistically equivalent to their target reading grade levels (p>0.05).

**2.5.2 Consistency of keywords and key phrases:** As noted in previous work (Huang et al., 2024a), retention of key concepts is a major obstacle in the task of textual simplification using LLMs, particularly when attempting to simplify text across a wider range of grade levels (12th to 4th grade). To measure the ability of the models to retain the most relevant keywords and phrases, we calculated two metrics. For keywords, we define **keyword accuracy** as the ratio of the correct keywords retained in the releveled passage ($KW_j$) to the set of keywords selected by the experienced human writers in the 12th Grade baseline passages ($KW_{Tot}$).

$$Keyword\ Accuracy = \frac{KW_j}{KW_{Tot}}$$

We use the **BERTScore** (Zhang et al., 2020) to calculate the **semantic similarity (consistency of the meaning of key phrases)** between the source sentences from the baseline 12th grade passages ($\mathscr{S} = s_1, s_2,..., s_n$) and the sentence regenerated by the LLM in the new leveled passage ($S^i = s_1^i, s_2^i,..., s_n^i$). Generally, a BERTScore closer to 1 is considered good, with scores between 0.70 and 0.85 being good to very good, and scores below 0.70 indicating lower similarity. This measurement is helpful because we can expect a lexical divergence between the two sentences but not a semantic one. BERTScore compares the semantic similarity of the source text and regenerated one by calculating the pairwise cosine similarities between pre-computed BERT (Devlin, J. 2019) token embeddings of each of the texts. BERTScores are conceptualized as an enhanced version of the more commonly known BLEU (Papineni et al. 2002) metric in natural language processing. While BERTScores are a newer metric, early evidence suggests that this metric may provide greater accuracy in capturing semantic similarity compared to BLEU (Hanna & Bojar, 2021).

**2.5.3 Word Count % Change:** As already noted by Huang and colleagues (2024a), LLMs can produce a shorter, paraphrased version of the original text, thus losing important details. For this reason, we measured the word count % change for each LLM/prompt combination, which measures the arithmetic mean change between the length of the source text and the regenerated text.

$$Word\ Count\ \%\ Change\ =\ \frac{1}{n}\sum_{i=1}^{n}\left|\left(L_{Source_i}\ -\ L_{RegenLLM_i}\right)\right|$$

We considered high word count consistency to be within the range $\pm 10\%$ ($\approx 30\ words$).

## 2.6 Procedure

The study was conducted to evaluate the ability of LLMs to simplify educational texts to specific grade levels while maintaining key content and structure. The following procedures were carried out in which to conduct the Experiment: 1) Develop 12th Grade Baseline Passage Set, 2) Model Selection of LLMs and 3) Conduct experiment, generating lower-level passages within each LLM and with the specified prompting techniques.

The experiment was conducted using publicly accessible datasets and models, with all prompting techniques and model configurations documented for reproducibility. For the calculation of BERTScore, we relied on the HuggingFace Evaluate library that makes evaluating and comparing models and reporting their performance easier and more standardized (Devlin et al., 2019). For other metrics including FKGL score, keyword and phrase accuracy, we developed an internal code available in our repository (Day et al., 2025). All datasets and code conducted in this work are publicly accessible via our repository for reproducibility (Day et al., 2025). The parameters used in these experiments, as well as the prompts can be found in the Supplemental Materials Section 3.

## 3. Results

We first manually reviewed all 2340 releveled passages for any anomalies before we analyzed the LLM/prompt combinations for their performance. We did find that for 25 releveled passages – eight from the grade 4 set, ten from the grade 6 set, and seven from the grade 8 set, all employing the Mixtral LLM and PC prompting technique – none of the text was relevant to the original passages. For example, some of the releveled passages contained "A:" or "Paragraph 1:" as the new text in its entirety. Other releveled entries contained what could be construed as a reiteration of the prompts given to the LLM; for example, "Sure, I can help you reduce the readability level of the source text from grade twelve to grade 8th, according to the Flesch-Kincaid Grade level. Here's the revised text:" In other words, Mixtral sometimes failed to simplify the source passage using prompt chaining. To ensure balanced comparisons of releveled texts across all LLM and prompting combinations, we opted not to include this text from the data set to be analyzed. Thus, 2315 releveled passages were included in the final dataset.

All statistical analyses were conducted in Minitab version 22 (Minitab LLC, 2024). To evaluate Grade Level Accuracy, we ran one-sample t-tests to compare the FKGL scores of the releveled passages and the targeted grade level for each prompt technique and LLM. For Keyword Accuracy, Semantic Similarity (BERTScores), and Word Count % Change, we ran multiple regression models to assess differences between prompt techniques and LLMs. Further, we also wanted to determine whether there were grade level differences in semantic similarity, thus we ran a linear regression model to assess whether semantic similarity (as determined by BERTScores) decreased as the gap between the targeted grade level of the releveled passages and the 12th grade baseline passage increased.

### 3.1 Grade Level Accuracy: Which LLM/prompting technique(s) had the best accuracy at downgrading passages to the targeted grade level?

Our first evaluation metric aimed to examine how closely each LLM and prompting method was able to match the targeted grade level for each modified passage. We first provide means (Table 2) for each targeted grade level and LLM/prompting method combination tested. One-sample t-tests were performed on the FKGL scores derived from each LLM/prompt combination's set of 60 releveled passages. These scores were compared to the midpoint within each grade level; for example, releveled fourth-grade passages' FKGL were compared to the midpoint between 4.00 - 4.99, or 4.50.

In Table 2, bolded averages indicate the releveled FKGL scores which were statistically equivalent to their target reading grade levels (p>0.05). When downgrading to the fourth grade, none of the LLM/prompt combinations yielded a significant mean FKGL's equivalent to the fourth-grade target of 4.5. At the sixth-grade level, GPT/DSP, Claude/Zero-shot, and Claude/COT combinations produced releveled passages equivalent to the sixth-grade target of 6.5. For eighth grade releveling, GPT/Zero-Shot, GPT/COT, GPT/Multi-Agent, and Mixtral/COT generated passages equivalent to the eighth-grade target of 8.5. Thus, the various LLMs and prompting methods tested were able to more accurately hit the desired grade level at 8th and 6th grade level but struggled when releveling from 12th grade down to 4th grade.

### 3.2 Keyword Accuracy, Semantic Similarity (BERTScores), and Word Count % Change

For Keyword Accuracy, Semantic Similarity (BERTScores), and Word Count % Change, we ran multiple regression models to determine whether there were significant differences in these metrics as determined by prompting technique and LLMs. We first provide a summary of means and standard deviations in Table 3 for the percentage of change in word count between the source text and the modified text, BERTScore (semantic similarity of key phrases), keyword accuracy between the 12th grade baseline passage and the releveled passages.

We also present bivariate correlations in Table 4 between BERTScore, Keyword Accuracy (percentage), Word Count Change %, and the releveled passage FKGL scores were tested for correlation significance with each other at the α = 0.05 level. BERTScore and keyword accuracy; BERTScore and FKGL; and keyword accuracy and FKGL were significantly positively correlated. BERTScore and percentage change in word count; and keyword accuracy and percentage change in word count; were all significantly negatively correlated. There was no significant correlation between percentage change in word count and FKGL (p=0.055).

### 3.3 Keyword Accuracy: Which LLM/prompting technique(s) had the highest consistency in maintaining keywords?

Keyword accuracy % indicates the consistency of the LLM/prompting technique combination to retain a given set of keywords in the releveled passage. A keyword accuracy of 100% means that the releveled text retained all of the keywords that were specified in the prompting technique. An initial model including both LLM and prompting technique as independent variables indicated that the LLM used had no significant bearing on keyword retention accuracy during passage releveling, thus were trimmed from the model. In a follow-up model including only prompting technique as the independent variable, DSP was the sole prompting technique that positively influenced keyword retention accuracy (Table 5). Zero-shot and CoT prompting negatively influenced keyword retention accuracy while multi-agent and PC had no significant impact on keyword retention.

### 3.4. Consistency of Key Phrases (semantic similarity): Which LLM/prompting technique(s) performed best at maintaining semantic similarity as measured by BERTScores across downgraded passages? Are there grade level differences such that semantic similarity decreases at lower grade levels?

BERTScore data reflect the ability of the LLM and prompting technique to retain the meaning of the original passage in the releveled passage. BERTScores were used then as a marker of consistency in key phrases between the 12th grade baseline passages and the lower leveled passages. Higher BERTScores are better (.85 and above indicate high similarity). The multiple regression model indicated that both LLM and prompting technique significantly predicted BERTScore. GPT-4 produced significantly higher BERTScores (higher similarity), while Mixtral tended to significantly negatively impact BERTScore (less similarity). As for the prompting techniques, DSP and CoT significantly predicted higher BERTScores, while the zero-shot and multi-agent prompting techniques significantly predicted declining BERTScores (Table 6).

As Farajidizaji et. al. (2024) found evidence of grade level differences in semantic similarity; we wanted to determine if there were grade level differences in BERTScores. Specifically, does semantic similarity decrease as the gap between the grade level of a source text and the modified text increases? In other words, do we observe higher semantic similarity when downgrading between a 12th grade and 8th grade passage compared to a 12th grade and 4th grade passage? To answer this question, we ran a linear regression model, choosing to use the BERTScores generated by the LLM and prompting method that appeared to maintain the highest level of semantic similarity, as noted above, GPT-4 LLM and the DSP prompting technique. Results confirmed that as the grade level gap between the original 12th grade passage and the releveled passages increased, semantic similarity decreased (Table 7; Figure 6). Pairwise Tukey comparisons of BERTScores at each grade level confirmed that the decreases were significant at the 95% confidence level.

### 3.5 Word Count % Change: Which LLM/prompting technique(s) performed best at maintaining word count?

As we observed, many LLMs tend to significantly reduce the length of a releveled passage, especially when attempting to make a larger decrease in grade level from the original source text. In the present study, smaller absolute percentage changes in word count are better. The multiple regression model indicated significant differences by both LLM and prompting technique (Table 8). Specifically, GPT-4 tended to lower the desired word count of approximately 300 words in the modified passages while Mixtral was associated with significantly increasing the word count. Claude 3 had no significant impact, which was most desirable. For prompting techniques, PC created a significant increase in word count percentage change and DSP drove a significant decrease (lowered word count). Neither zero-shot, Multi-Agent nor CoT had a significant impact on word count percentage change, thus were best at maintaining desired word count.

## 4. Discussion

### 4.1 General Findings

The primary aim of this exploratory study was to evaluate the use of Large Language Models (LLMs) and novel prompting techniques for the purpose of text simplification of educational content. Previous research has well-documented the importance of providing developing or struggling readers with content that is matched with their current level of reading skills (Alowais, 2021; Denton et al., 2014). When struggling readers are provided with texts well above their skill level, reading fluency, comprehension, and motivation are negatively impacted (Amendum et al., 2017). General education classrooms typically include students with a wide range of reading skills. Thus, while being able to provide students with more individualized learning materials may foster higher learning outcomes, teachers are not commonly provided with multiple versions of their curriculum at various grade levels. Further, text simplification is complex and requires significant time and training which are typically beyond what most educators have available. GenAI carries great potential in which to build effective tools that can assist teachers in automatically releveling texts to provide differentiated instruction to students. However, previous work has noted challenges in LLMs' ability to accurately relevel and maintain consistency (Farajidizaji et. al. 2024; Huang et al., 2024a; Patel et al., 2022).

In the current experiment, we aimed to systematically compare the use of three LLMs (GPT-4 Turbo, Claude 3, and Mixtral 8x22B) and three known prompt engineering techniques (Chain-of-Thought prompting, Prompt Chaining, and Directional Stimulus Prompting (DSP)), and tested a novel multi-agent architecture in lowering the grade level of a baseline set of human-written passages covering a variety of informational topics (science, social studies, humanities, etc.) leveled at 12th grade down to 8th grade, 6th grade, and 4th grade. We assessed metrics of accuracy of matching the desired grade level of the releveled passages, and consistency of keywords, phrases (semantic similarity), and word count. While we acknowledge there are other metrics that could be relevant to consider, we chose to focus on these four as they have been identified as challenges in previous research (Farajidizaji et. al. 2024; Huang et al., 2024a; Patel et al., 2022). We provide a summary table of the results in Table 9.

First, we evaluated which LLM and prompting technique achieved the greatest accuracy in hitting the desired grade level for each modified passage. At the 8th and 6th grade levels, we found that all LLMs and multiple prompting techniques performed relatively well at modifying the 12th grade baseline passages to the targeted grade level. GPT-4 performed well at reaching both 8th and 6th grade target levels. When releveling down to the fourth-grade level, however, almost all of the LLMs and prompt techniques generated passage means above the 4th grade level. Of the LLM/prompt techniques combinations that were tested, Claude 3 with Prompt Chaining produced the closest match at a mean grade level of 3.45. These results are not particularly surprising and replicate findings from Farajidizaji et al. (2024). LLMs appear to have difficulty in generating texts of a lower readability level and these differences may be even more apparent when attempting to simplify down from upper grade levels. If hitting an exact level of difficulty is not of particular importance, a simplified passage that is slightly off the target may not matter. Educators or others attempting to use LLMs to modify the grade level of texts should exercise caution in reviewing modified content to ensure it is appropriate and may wish to consider verifying the grade level of the modified text using readability calculators such as Coh-Metrix or Lexile.

For consistency in maintaining keywords, while we found no significant differences in LLMs, we did find that directional stimulus prompting (DSP) maintained keywords across grade levels with the greatest accuracy. One of the primary aims of this work was to assess how we could improve keyword consistency through various prompting techniques as learning specific vocabulary may be of particular importance in educational contexts, for example, when students will be taking a test that will assess their knowledge and comprehension of specific vocabulary and/or details on a topic. It is not surprising that DSP produced the best keyword consistency because, unlike PC and Multi-Agent, the human-annotated keywords served to guide the LLM.

We also wanted to consider that in addition to specific words, an educator might want to ensure consistency of concepts or ideas on a more general level across simplified versions of the text. We chose to assess key phrase accuracy using BERTScores (Zhang et al., 2020). Our results suggested that for LLMs, Claude 3 and GPT-4 both demonstrated high consistency in maintaining semantic similarity across simplified passages while Mixtral produced significantly lower consistency. We also found main effects for DSP and CoT prompting techniques that produced significantly higher consistency in key phrases. Again, as DSP utilized key phrases directly chosen by our expert writers, this finding is logical. In the case of CoT, it is likely dependent on the reasoning chain and subdivision of the problem into sub-tasks that improve reasoning skills. Further, given the complexity of maintaining semantic similarity, we wanted to investigate whether there were grade level differences such that the further away in grade level the modified passages got from the 12th grade baseline passage, that semantic similarity would decrease. Replicating results from Farajidizaji et. al. (2024), as the gap increases between the source text and the simplified text, LLMs and prompting techniques struggled to maintain semantic similarity. However, it is worth acknowledging that the BERTScore metric is relatively new, and more research will be needed to determine its accuracy in measuring semantic similarity (Hanna and Bojar, 2021). Thus, particular caution may be required if attempting to simplify texts across a wider range of grade levels.

Finally, it is also important to note that while we used BERTScores as a metric to capture consistency in key phrases, the scores were calculated considering the entire passage so while we aimed to maintain key phrases through our prompting techniques, these scores capture similarity of the entire passage. We made an assumption that higher BERTScores likely also captured key phrases, but this may not be entirely accurate and there may be other metrics to consider to better capture consistency of specific details. Nonetheless, Directional Stimulus Prompting was the most effective prompting method for maintaining both keywords and key phrases.

The last metric we evaluated was the degree to which each LLM and prompting method reduced the number of words from the baseline 12th grade passages that were all approximately 300 words. While maintaining a specific number of words may not always be of particular importance in the task of text simplification, we have observed that many LLMs and existing tools significantly cut text when simplifying text. Indeed, this problem had been observed before (Huang et al., 2024a). Significant loss of words in a simplified text likely indicates that LLMs were paraphrasing the original version of the text, which is not ideal for the task at hand. In general, Claude 3 appeared to be the LLM model that best maintained the original length of the text. For prompting techniques, zero-shot, multi-agent, and CoT were most effective in maintaining length.

As previous research has found shortcomings in LLMs and prompting techniques, we also tested the use of a novel Multi-agent architecture. While the multi-agent architecture was less effective for three of the tested metrics, it did excel in maintaining length, even when leveling down to the 4th grade level. This is due to the iterative nature of the method, where each generated text is evaluated (Editor) and rewritten (Writer) from time to time to achieve the right trade-off between minimum length and target FKGL score. The model, therefore, sometimes makes choices that necessarily compromise semantic and lexical similarity from the original text. For example, rewriting sentences that are not considered relevant or substituting some key words. In addition, keywords consistency is marred by the fact that the selection is at the discretion of the LLM and not suggested by human editors as in DSP. This approach, therefore, slightly reduces the similarity from the source text in favor of the performance of the task. It is important to note that this workflow could only be tested with the OpenAI models (GPT-4) at this time.

**4.3 Results Summary**

 Taken together, for LLMs, GPT-4 performed well at reaching grade level accuracy at 8th and 6th grade and was also strongest in maintaining key phrases. Directional Stimulus Prompting as a technique was effective in maintaining both keywords and semantic similarity. Chain of Thought prompting was also effective in maintaining semantic similarity and also performed well in maintaining passage length (word count). For grade level accuracy, all LLMs and prompting techniques performed relatively well at matching the desired grade level for 6th and 8th grade, but struggled when modifying down to 4th grade. Similarly, we also observed a significant decline in semantic similarity of key phrases when leveling passages down from 12th to 4th grade. Overall, careful consideration should be given as to the specific LLM or prompting technique to be utilized, depending on the specific task and context. Further, avoiding attempting to make larger leaps between grade levels (i.e. 12th grade to 4th grade vs. 8th grade to 6th grade) may also help to improve the accuracy and consistency of the simplified text.

**4.3 Limitations and Directions for Future Research**

 There were a few limitations worth noting in the current study. While we primarily focused on accuracy of grade level and passage consistency in releveled passages, we did not thoroughly evaluate the overall quality of the releveled passages. While it could be argued that indirectly, scores like BERTScore could be indicators of quality when consistency is higher, it would first require the assumption that the baseline texts are of high quality based on certain metrics/validation. Thus, consistency does not necessarily equate to high quality. In the context of text simplification, concerns of quality may be of less focus if the source text is of high quality. While improved prompting techniques can potentially reduce quality issues, more research will be needed to assess the quality of AI-generated content for use in education. Our future work includes developing a rubric in which AI-generated content can be evaluated for quality and considers factors such as repetitions, cohesion, burstiness, spelling, and grammatical errors. Future work may also consider incorporating additional readability metrics.

 We opted to use 300-word passages in the current study. We aimed to achieve a balance between using shorter passages which could have been too restrictive and yet long enough that there were sufficient details to detect differences in the output produced by the LLMs and prompting techniques. We initially tried this task with passages longer than 600 words and observed a tendency in LLMs to generate increasingly shorter passages (less than 500 words), a pattern also noted by Huang and colleagues (2024a). Thus, text simplification with longer texts may be a limitation of the current capabilities of LLMs. Future studies should consider using passages at varying lengths to evaluate differences in accuracy and consistency as LLMs are updated over time. Additionally, using larger datasets of passages will also be important for better understanding LLM performance.

It is also worth noting that the number of keywords and key phrases identified in the 12th grade baseline passages by our human content writers were sometimes quite varied depending on the evaluator and topic of the passages. Although it was beyond the scope of this initial study, it is possible that differences in the number of keywords and phrases selected could impact keyword and key phrase (semantic similarity) consistency across each simplified passage. An important next step in this work would be to determine potential thresholds across LLMs in which consistency may be negatively impacted or in which an LLM is unable to carry out the task of modifying a text. Thresholds may also vary depending on the length of a text or even the topic (Huang et al., 2024a).

While future work may continue to consider a Multi-Agent workflow, it is worth noting the main limitation in AutoGen stems from its high cost and slow inference time. Costs are driven by how the framework handles context, distributing the entire conversation to all agents, resulting in an average cost of $2.43 for 11 requests and 21,500 tokens. The recursive workflow structure delays results, averaging over 4 minutes. Additionally, issues were noted in agent selection and understanding, with agents like the Editor sometimes struggling to grasp tasks. Biases in prompting techniques led to text shortening, illogical keyword choices, and difficulty paraphrasing complex content, risking meaning distortion and hallucinations. Future work should consider additional applications and metrics; for example, Prometheus, a 13B evaluator LLM that can assess any given long-form text based on a customized score rubric provided by the user (Kim et al., 2024). Finally, additional techniques and LLMs (such as fine-tuned LLMs) could be explored as these models may have greater accuracy and can be integrated into either a multi-agent workflow or a Retrieval-Augmented Generation (RAG) pipeline which could help reduce content loss and hallucinations (Shuster et al., 2021). Finally, in addition to content modifications, LLMs could be leveraged to enhance readability through visual changes to the text, such as increasing font size or letter spacing to further support fluency and comprehension (Day et al., 2024; Beier et al., 2022).

**5. Conclusion**

This exploratory study provides a foundational approach and suggested metrics for future research into the application of LLMs for the purpose of automated text simplification of educational texts and is one of the most rigorous evaluations to date. Through the use of advanced prompting techniques such as Prompt Chaining, Chain-of-Thought, and Directional Stimulus Prompting, alongside a novel multi-agent framework, we evaluated how effectively LLMs could modify texts to targeted grade levels while retaining key information and word count. For grade level accuracy, we observed greater accuracy across multiple LLMs and prompting techniques for passages modified down to the 8th and 6th grade level. However, when leveling down to 4th grade, all the LLMs and techniques struggled to accurately hit the grade level targets. Key phrase and keyword retention improved through Directional Stimulus Prompting, suggesting that this method holds promise for enhancing the consistency and quality of educational materials. Chain-of-Thought prompting was also effective at maintaining consistency of key phrases (semantic similarity) and word count, avoiding a common LLM pitfall of attempting to paraphrase text.

However, significant challenges remain in balancing the consistency of keywords and main ideas when reducing text difficulty. Careful consideration should be given to determine which models and techniques might be most effective depending on the goals of the task. Further, GenAI can only achieve its potential if it can behave in ways that engender educator trust by providing contextually appropriate output. Models are continually being updated, and they will require iterative evaluation to understand how performance evolves over time. The present study provides a generalizable method to assess future systems. Including educators in the development of new methods and tools through design-based implementation research can also ensure that they are most effective and appropriate for classroom use. Future research should include texts on varied lengths and topics and consider additional metrics for evaluating the quality of LLM-generated content. The results from this study suggest that GenAI holds great promise for supporting differentiated instruction by automating the complex and time-consuming process of text simplification, providing educators with potentially more efficient simplification methods and tools and ultimately, enhancing student learning outcomes.

**Tables**

**Table 1**

*Sample Keywords and Key Phrases from The American Hippo 12th Grade Passage*

| Keywords | Key Phrases |
|---|---|
| American Hippo Bill; Louisiana; bayous; meat shortage; and lake cow bacon | importation and release of hippopotamuses; eat the invasive water hyacinth; produce meat; Cities were expanding rapidly; turning former "wasted" bayou land into farming opportunity; full pockets; full bellies; Chicago slaughterhouses; local meatpacking industry; and it was never passed |

**Table 2**

*Means and Standard Deviations of FKGL Scores for Releveled Passages by LLM/Prompt*

| | Target: 4th Grade (FKGL = 4.5) | | | Target: 6th Grade (FKGL = 6.5) | | | Target: 8th Grade (FKGL = 8.5) | | |
|---|---|---|---|---|---|---|---|---|---|
| LLM → Prompt | GPT 4 | Mixtral 22X8B | Claude 3 | GPT 4 | Mixtral 22X8B | Claude 3 | GPT 4 | Mixtral 22X8B | Claude 3 |
| Zero-shot | 6.88*** (1.20) | 6.29*** (1.48) | 5.36*** (1.43) | 7.66*** (1.13) | 6.92* (1.45) | **6.88** (1.58) | **8.40** (1.26) | 7.78*** (1.39) | 7.87** (1.49) |
| PC | 5.93*** (1.26) | 7.24*** (2.56) | 3.45*** (0.80) | 6.91* (1.52) | 8.44*** (2.05) | 6.05* (1.28) | 8.00* (1.62) | 9.47** (2.04) | 6.16*** (1.09) |
| DSP | 5.95*** (1.13) | 7.97*** (1.91) | 6.08*** (1.52) | **6.61** (1.53) | 8.19*** (1.93) | 7.84*** (1.77) | 7.98* (1.79) | 10.04*** (1.74) | 9.90*** (1.56) |
| CoT | 6.09*** (1.30) | 6.24*** (1.61) | 5.24** (1.63) | 6.84* (1.27) | 7.19** (2.02) | **6.56** (1.66) | **8.35** (1.52) | **8.78** (1.67) | 7.43*** (1.72) |
| Multi-Agent (GPT4 only) | 5.31*** (0.95) | | | 6.73** (0.53) | | | **8.49** (0.045) | | |

*Note.* *p<0.05; **p<0.01; ***p<0.001. Bolded averages indicate the releveled FKGL scores which were statistically equivalent to their target reading grade levels.

**Table 3**

*Means and Standard Deviations for Word Count Change %, Semantic Similarity (BERTScores), and Keyword Accuracy as a Function of LLM/Prompt Combination*

| | Word Count % Change | | | BERTScore | | | Keyword Accuracy (%) | | |
|---|---|---|---|---|---|---|---|---|---|
| **LLM →**<br><br>**Prompt** | **GPT 4** | **Mixtral 8x22B** | **Claude 3** | **GPT 4** | **Mixtral 8X22B** | **Claude 3** | **GPT 4** | **Mixtral 8x22B** | **Claude 3** |
| Zero-shot | 11.33 (10.43) | 22.66 (15.39) | 44.55 (22.23) | 0.898 (0.015) | 0.905 (0.015) | 0.907 (0.023) | 55.70 (21.31) | 59.23 (20.49) | 62.73 (22.63) |
| PC | 11.76 (12.14) | 216.32 (554.21) | 13.03 (10.07) | 0.932 (0.027) | 0.905 (0.033) | 0.915 (0.023) | 70.81 (20.76) | 62.27 (27.89) | 59.09 (23.75) |
| DSP | 16.65 (13.46) | 18.34 (14.65) | 20.38 (16.32) | 0.922 (0.027) | 0.929 (0.021) | 0.939 (0.026) | 73.28 (17.77) | 77.88 (17.77) | 80.88 (17.86) |
| CoT | 12.87 (11.73) | 28.46 (36.84) | 35.82 (26.48) | 0.937 (0.025) | 0.921 (0.035) | 0.919 (0.025) | 63.82 (21.77) | 66.17 (24.27) | 57.86 (22.42) |
| Multi-Agent (GPT4 only) | 8.68 (8.26) | | | 0.911 (0.034) | | | 65.74 (23.90) | | |

**Table 4**

*Correlations of BERTScores, Releveled FKRA, Keyword Accuracy, and Word Count Change %*

|  | BERTScore | Accuracy Keywords | % Change N Words |
|---|---|---|---|
| Accuracy Keywords | 0.505*** |  |  |
| % Change N Words | -0.301*** | -0.190*** |  |
| Releveled FKRA | 0.484*** | 0.267*** | -0.040 |

***p<0.001

**Table 5**

*Multiple Regression Model of Prompting Technique Used to Maintain Keyword Accuracy %*

| Term | Coef | SE Coef | T-Value | P-Value |
|---|---|---|---|---|
| Constant | 65.81 | 0.51 | 129.18 | <0.001 |
| Prompt | | | | |
| COT | -3.20 | 0.91 | -3.55 | <0.001 |
| DSP | 11.54 | 0.91 | 12.80 | <0.001 |
| Multi-Agent | -0.08 | 1.38 | -0.05 | 0.956 |
| PC | -1.67 | 0.92 | -1.82 | 0.068 |
| Zero-shot | -6.59 | 0.90 | -7.32 | <0.001 |

$R^2$=8.18%

**Table 6**

*Multiple Regression Model of LLM and Prompting Technique Used to Retain Semantic Similarity (BERTScore)*

| Term | Coef | SE Coef | T-Value | P-Value |
| --- | --- | --- | --- | --- |
| Constant | 0.917 | 0.001 | 1448.51 | <0.001 |
| LLM | | | | |
| Claude 3 | 0.001 | 0.001 | 1.05 | 0.295 |
| GPT-4 | 0.003 | 0.001 | 3.67 | <0.001 |
| Mixtral 8x22B | -0.003 | 0.001 | -4.68 | <0.001 |
| Prompt | | | | |
| COT | 0.008 | 0.001 | 7.78 | <0.001 |
| DSP | 0.013 | 0.001 | 12.05 | <0.001 |
| Multi-Agent | -0.009 | 0.002 | -4.89 | <0.001 |
| PC | 0.001 | 0.001 | 0.81 | 0.417 |
| Zero-shot | -0.014 | 0.001 | -12.68 | <0.001 |

$R^2$=13.36%

**Table 7**

*Linear Regression Model of GPT-4/DSP Used to Maintain Semantic Similarity by Grade Level*

| Term | Mean | Coef | SE Coef | T-Value | P-Value |
|---|---|---|---|---|---|
| Constant | | 0.922 | 0.001 | 536.96 | <0.001 |
| Grade Level | | | | | |
| 4 | 0.908 | -0.014 | 0.002 | -5.94 | <0.001 |
| 6 | 0.919 | -0.004 | 0.002 | -1.52 | 0.131 |
| 8 | 0.941 | 0.018 | 0.002 | 7.45 | <0.001 |

$R^2=25.97\%$

**Table 8**

*Multiple Regression Model of LLM and Prompting Technique Used to Maintain Word Count*

| Term | Coef | SE Coef | T-Value | P-Value |
|---|---|---|---|---|
| Constant | 35.51 | 3.54 | 10.03 | <0.001 |
| LLM | | | | |
| Claude 3 | -7.96 | 4.56 | -1.75 | 0.081 |
| GPT-4 | -23.26 | 4.56 | -5.10 | <0.001 |
| Mixtral 8x22B | 32.22 | 4.60 | 6.78 | <0.001 |
| Prompt | | | | |
| COT | -9.80 | 6.11 | -1.60 | 0.109 |
| DSP | -17.06 | 6.11 | -2.79 | 0.005 |
| Multi-Agent | 3.58 | 9.96 | -0.36 | 0.720 |
| PC | 39.77 | 6.20 | 6.42 | <0.001 |
| Zero-shot | -9.34 | 6.11 | -1.53 | 0.127 |

$R^2$=4.17%

**Table 9**

*Results Summary Table of Best Performing LLM(s) and Prompt Technique(s)*

| Metric | Best Performing LLM(s) | Best Performing Prompt Technique(s) | 12th Grade Passage Sample Section | Simplified Version Using Best LLM and Prompting Technique |
|---|---|---|---|---|
| Grade Level Accuracy | GPT-4 & Mixtral 8x22B | Chain-of-Thought & Prompt Chaining | The issue of sleeping is of considerable importance to the National Basketball Association (NBA), as traveling and packed game schedules are among the aspects of the NBA calendar that affect the sleep of NBA personnel. Due to these and other factors, sleep deprivation has become a prevalent issue affecting player performance. To help combat sleep deprivation, NBA organizations have employed scientists or doctors specializing in sleep and sleep medicine on their staff. | The issue of sleeping is very important to the NBA. Traveling and busy game schedules are parts of the NBA calendar that affect the sleep of NBA people. Because of this and other factors, lack of sleep has become a common problem affecting player performance. To help fight lack of sleep, NBA teams have hired scientists or doctors who specialize in sleep. *(6th Grade)* |
| Keyword Accuracy | No Significant Differences | Directional Stimulus Prompting | Secretariat, also known as Big Red, was a champion American thoroughbred racehorse who was the ninth winner of the American Triple Crown, setting and still holding the record for fastest time. He is considered to be the greatest racehorse of all time. Secretariat became the first Triple Crown winner in 25 years, and his record-breaking victory in the Belmont Stakes, which he won by thirty-one lengths, is regarded as one of the greatest races ever. | Secretariat was a champion racehorse. He was also called Big Red. He won the American Triple Crown. He set records for the fastest time. Many think he was the greatest racehorse ever. Secretariat was the first Triple Crown winner in 25 years. His win at the Belmont Stakes was by 31 lengths. It is seen as one of the greatest races. *(4th Grade)* |
| | Claude 3 & GPT-4 | Directional Stimulus Prompting & Chain-of-Thought | Orcas are the most formidable predators to roam the ocean. Their superior intelligence, creativity, cooperation, and physical bodies make them highly effective, and they use several ingenious hunting techniques, including the "wave wash" and "karate chop." The "wave wash" is a seal hunting technique devised to counter defensive perching on ice floes in which a group of orcas organize into a tight formation, charge the ice floe, causing a large wavefront, and give it one last push with their tails as they swim underneath. | Orcas are the strongest hunters in the ocean. Their smarts, creativity, teamwork, and strong bodies make them great hunters. They use clever hunting tricks like the "wave wash" and "karate chop." The "wave wash" is a trick to catch seals. A group of orcas swim close together, charge an ice floe, make a big wave, and push it with their tails to flip the seal into the water. *(6th Grade)* |
| Word Count Change | Claude 3 | Multi-Agent, Zero-Shot, & Chain-of-Thought | A black hole is a region of spacetime where gravity is so powerful that nothing, including light or other electromagnetic waves, has sufficient energy to escape. The presence of a black hole can be inferred through its interaction with other matter and radiation, including visible light. Any matter that falls onto a black hole becomes heated by friction, forming quasars, some of the brightest objects in the universe. Stars passing too close to a supermassive black hole can be shredded and swallowed. | A black hole is a region of spacetime where gravity is so powerful that nothing, including light or other electric rays, has enough energy to escape. The presence of a black hole can be seen through its effect on other matter and rays, including visible light. Any matter that falls onto a black hole becomes heated by friction. This forms bright objects, some of the brightest in the universe. Stars passing too close to a supermassive black hole can be shredded and swallowed. *(8th Grade)* |

**Figures**

**Figure 1**

*Directional Stimulus Prompting (DSP) Prompt Structure*

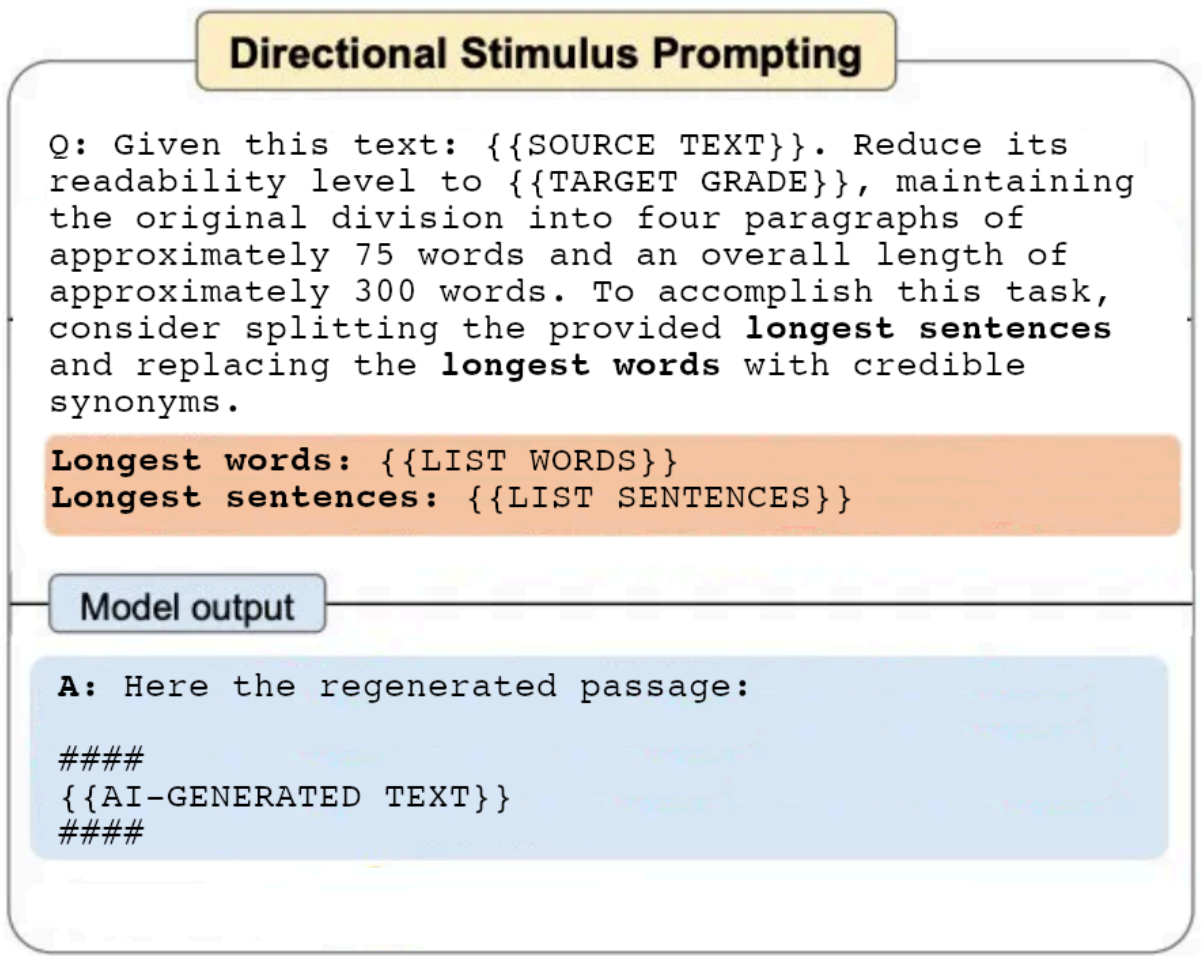

**Figure 2**

*Automatic Chain-of-Thought Prompting Examples*

**TWO EXAMPLES**

> **Q: Given this text: {{SOURCE-TEXT #1}}. Reduce its readability level to {{TARGET GRADE}},maintaining the original division into 4 paragraphs of approximately 75 words and an overall length of approximately 300 words.**
>
> **A: Let's think step by step. The source text has several long sentences: {{SENTENCES #1}}. These could be splitted like that: {{SPLITTED SENTENCES #1}}. Furthermore, we identify these words as the longest (3+ syllables): {{WORDS #1}}. We can replace them with these synonyms: {{ SYNONYMS #1}}**
>
> **Here the regenerated text: {{AI-GENERATED TEXT #1}}**
>
> **Q: Given this text: {{SOURCE-TEXT #2}}. Reduce its readability level to {{TARGET GRADE}}, maintaining the original division into 4 paragraphs of approximately 75 words and an overall length of approximately 300 words.**
>
> **A: Let's think step by step. The source text has several long sentences: {{SENTENCES #2}}. These could be splitted like that: {{SPLITTED SENTENCES #2}}. Furthermore, we identify these words as the longest (3+ syllables): {{WORDS #2}}. We can replace them with these synonyms: {{SYNONYMS #2}}**
>
> **Here the regenerated text: {{AI-GENERATED TEXT #2}}**

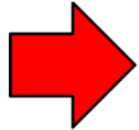

**QUESTION**

> Q: Given this text: {{SOURCE-TEXT}}
>
> Reduce its readability level to **{{TARGET GRADE}}**, maintaining the original division into 4 paragraphs of approximately 75 words and an overall length of approximately 300 words.
>
> A: Let's think step by step.

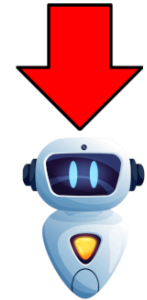

**LLM**

**Figure 3**

*Workflow of Prompt Chaining (PC)*

### SOURCE TEXT

House Resolution 23261, also known as the "American Hippo Bill," was a bill introduced by Representative Robert F. Broussard of Louisiana in 1910 to authorize the importation and release of hippopotamuses into the bayous of Louisiana. His objective was twofold. He hoped the hippos would eat the invasive water hyacinth that was clogging the rivers, solving an ecological problem, and also produce meat to help solve the American meat predicament, solving an economic problem.

There was a meat shortage because the meat industry couldn't keep up with high immigration levels. Cities were expanding rapidly, and Americans needed more meat. Although the transcontinental railroad was in place, it would take many more years to develop the cattle ranches we now think are commonplace in the American west. Politicians needed solutions, so the hippo plan was hatched.
[...]

**FKRA:** 12.4

### KEYWORDS AND KEYPHRASES

####
Keywords
####
- American Hippo Bill
- Louisiana
- bayous
- meat shortage
- lake cow bacon

####
Key Phrases
####
- importation and release of hippopotamuses
- eat the invasive water hyacinth
- produce meat
- Cities were expanding rapidly
- turning former "waster" bayou land into farming opportunity

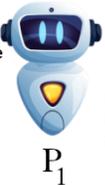

$P_1$

### LEVELED-TEXT GENERATION

House Resolution 23261 was also known as the "American Hippo Bill." The bill was introduced by Rep, Robert F. Broussard of Louisiana in 1910. He sought to authorize the import and release of hippopotamuses into bayous. He had two goals. He hoped the hippos would eat the invasive water hyacinth clogging the rivers. He also wanted to produce meat to help solve the meat shortage. This would solve both ecological and economic problems.

The meat industry just couldn't keep up with immigration. Cities were growing quickly. People needed more meat. The transcontinental railroad was in place. Still, it would take many more years to develop the cattle ranches that are now common in the West. Politicians needed solutions. As a result, the hippo plan was hatched. Louisiana was the perfect place for hippos to graze. They would turn former "wasted" bayou land into farmland, full pockets, and full bellies.

**FKRA:** 6.3

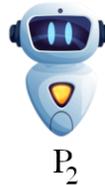

$P_2$

**Figure 4**

*Multi-Agent Method Architecture*

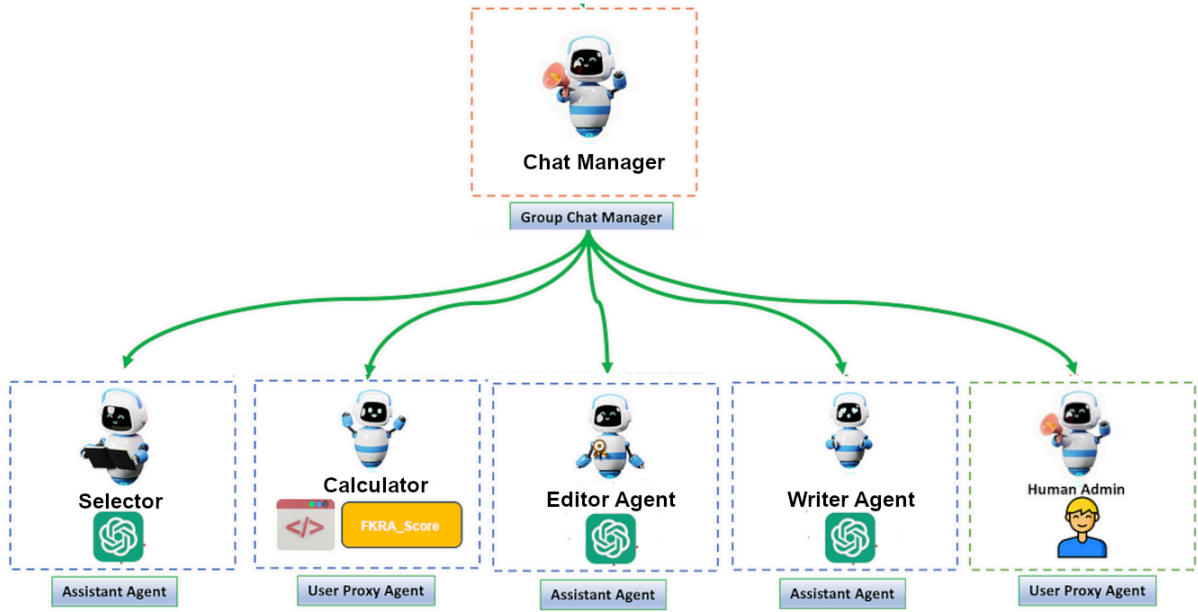

**Figure 5**

*Diagram of the Grade-Leveled Passage Generation Task*

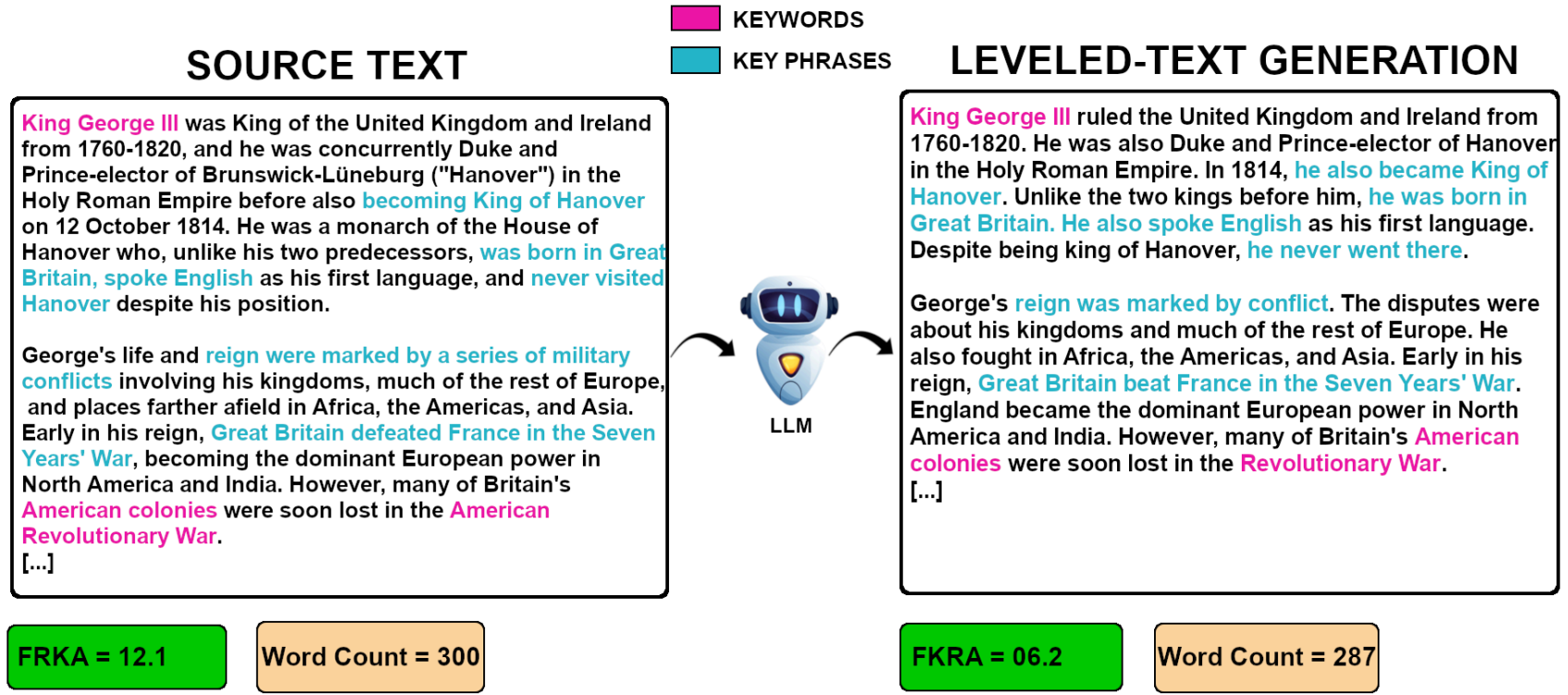

**Figure 6**

*BertScores by Grade Level*

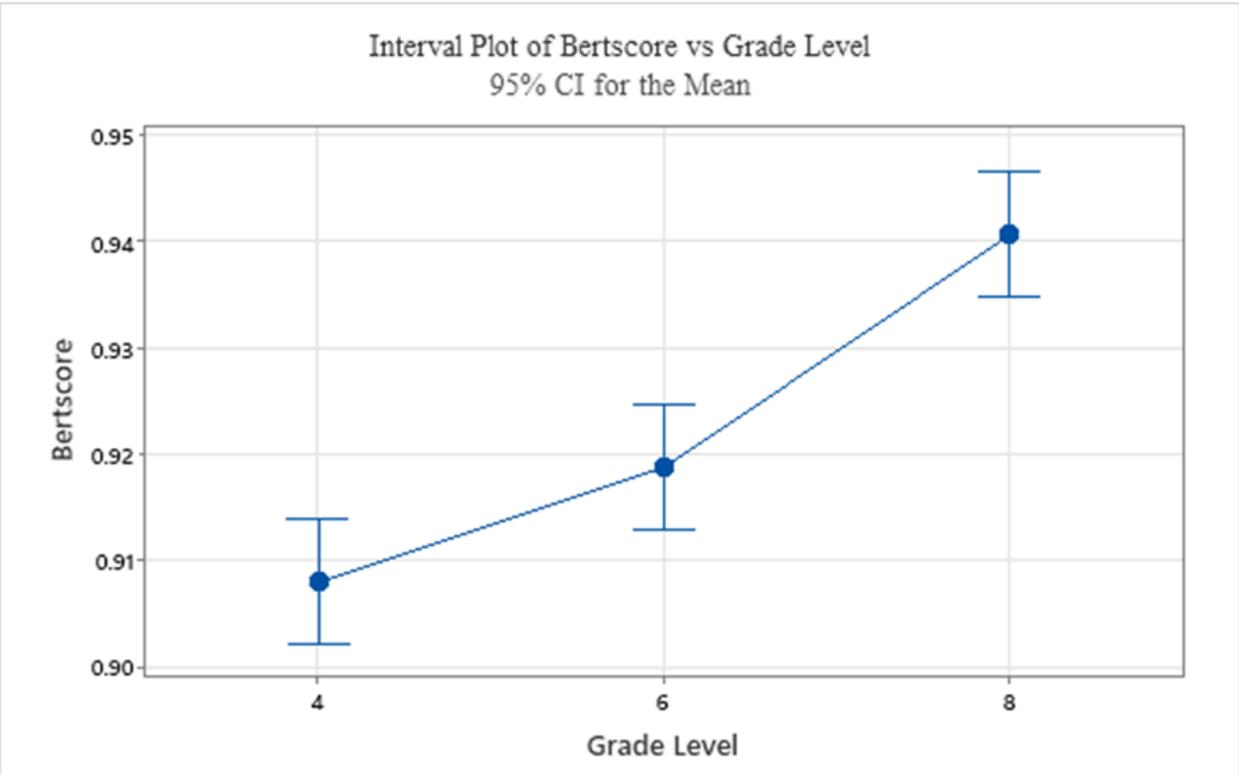

*Note.* The pooled standard deviation is used to calculate the intervals.

## References

Alowais, A. (2021). The Effects of Leveled Reading on Second Language Learners. International Journal of Research in Education and Science, 7(4), 1281-1299.

Alonzo, O., Seita, M., Glasser, A., & Huenerfauth, M. (2020). Automatic text simplification tools for deaf and hard of hearing adults: Benefits of lexical simplification and providing users with autonomy. In Proceedings of the 2020 CHI Conference on Human Factors in Computing Systems (pp. 1–13).

Amendum, S. J., Conradi, K., & Hiebert, E. (2017). Does Text Complexity Matter in the Elementary Grades? A Research Synthesis of Text Difficulty and Elementary Students' Reading Fluency and Comprehension. Educational Psychology Review, 30(1), 121–151. https://doi.org/10.1007/s10648-017-9398-2

Anthropic. (n.d.). *Getting started.* Retrieved from https://docs.anthropic.com/en/api/getting-started. Accessed December 15, 2024.

Barzilay, R., & Elhadad, N. (2003). Sentence Alignment for Monolingual Comparable Corpora. Conference on Empirical Methods in Natural Language Processing.

Beier, S., Berlow, S., Boucaud, E., Bylinskii, Z., Cai, T., Cohn, J., Crowley, K., Day, S. L., Dingler, T., Dobres, J., Healey, J., Jain, R., Jordan, M., Kerr, B., Li, Q., Miller, D. B., Nobles, S., Papoutsaki, A., Qian, J., Rezvanian, T., Rodrigo, S., Sawyer, B. D., Sheppard, S. M., Stein, B., Treitman, R., Vanek, J., Wallace, S., & Wolfe, B. (2022). Readability research: An interdisciplinary approach. *Foundations and Trends® in Human–Computer Interaction, 16*(4), 214–324.

Calderón, M., Slavin, R., & Sanchez, M. (2011). Effective instruction for English learners. The future of children, 103-127.

Chen, X., Xie, H., Zou, D., & Hwang, G.-J. (2020). Application and theory gaps during the rise of Artificial Intelligence in Education. Computers and Education: Artificial Intelligence, 1, 100002. https://doi.org/10.1016/j.caeai.2020.100002

Connor, C. M., Morrison, F. J., Fishman, B., Crowe, E. C., Al Otaiba, S., & Schatschneider, C. (2013). A longitudinal cluster-randomized controlled study on the accumulating effects of individualized literacy instruction on students' reading from first through third grade. Psychological Science, 24(8), 1408-1419.

Crossley, S. A., Allen, D., & McNamara, D. S. (2011). Text simplification and comprehensible input: A case for an intuitive approach. Language Teaching Research, 16(1), 89–108. doi:10.1177/1362168811423456

Day, S.L., Cirica, J., Clapp, S.R., Penkova, V., Giroux, A.E., Bordeau, C., Banta, A., Mutteneni, P., & Sawyer, B.D. (2025). Virtual Readability Lab: LLM. Github. https://github.com/virtual-readability-lab/LLM

Day, S. L., Atilgan, N., Giroux, A. E., & Sawyer, B. D. (2024). The Influence of Format Readability on Children's Reading Speed and Comprehension. *Education Sciences*, *14*(8), 854. https://doi.org/10.3390/educsci14080854

Denton, C. A., Fletcher, J. M., Taylor, W. P., Barth, A. E., & Vaughn, S. (2014). An experimental evaluation of guided reading and explicit interventions for primary-grade students at-risk for reading difficulties. Journal of Research on Educational Effectiveness, 7(3), 268-293.

Devlin, J., Chang, M.-W., Lee, K., & Toutanova, K. (2019). BERT: Pre-training of Deep Bidirectional Transformers for Language Understanding. Proceedings of the 2019 Conference of the North, 1. https://doi.org/10.18653/v1/n19-1423

DuBay, W. H. (2007). Smart Language: Readers, Readability, and the Grading of Text. In ERIC. https://eric.ed.gov/?id=ED506403

Dubois, Y., Galambosi, B., Liang, P., & Hashimoto, T. B. (2024). Length-controlled AlpacaEval: A simple way to debias automatic evaluators. doi:10.48550/ARXIV.2404.04475

Evans, R., Orăsan, C., & Dornescu, I. (2014). An evaluation of syntactic simplification rules for people with autism (pp. 131–140). Association for Computational Linguistics. https://aclanthology.org/W14-1215.pdf

Farajidizaji, A., Raina, V., & Gales, M. (2024, May 27). Is it Possible to Modify Text to a Target Readability Level? An Initial Investigation Using Zero-Shot Large Language Models. ArXiv.org. https://doi.org/10.48550/arXiv.2309.12551

Feng, Y., Qiang, J., Li, Y., Yuan, Y., & Zhu, Y. (2023). *Sentence Simplification via Large Language Models*. arXiv. https://arxiv.org/abs/2302.11957

Firmender, J. M., Reis, S. M., & Sweeny, S. M. (2013). Reading comprehension and fluency levels ranges across diverse classrooms: The need for differentiated reading instruction and content. Gifted Child Quarterly, 57(1), 3-14.

Flesch, R. (1948). A new readability yardstick. J*ournal of Applied Psychology*, 32(3), 221–233. https://doi.org/10.1037/h0057532

García-Peñalvo, Francisco & Vázquez-Ingelmo, Andrea. (2023). What Do We Mean by GenAI? A Systematic Mapping of The Evolution, Trends, and Techniques Involved in Generative AI. International Journal of Interactive Multimedia and Artificial Intelligence. 8. 7-16. 10.9781/ijimai.2023.07.006.

Hanna, M., & Bojar. (2021). A Fine-Grained Analysis of BERTScore. In *Proceedings of the Sixth Conference on Machine Translation*, pages 507–517, Online. Association for Computational Linguistics.

Huang, C. Y., Wei, J., & Huang, T. H. K. (2024a). Generating Educational Materials with Different Levels of Readability using LLMs. *In2Writing 2024.*

Huang, X., Liu, W., Chen, X., Wang, X., Wang, H., Lian, D., Wang, Y., Tang, R., & Chen, E. (2024b). Understanding the planning of LLM agents: A survey. *arXiv*. https://arxiv.org/abs/2402.02716

HuggingFace. (n.d.). *Mixtral-8x22B-v0.1.* Retrieved from https://huggingface.co/mistralai/Mixtral-8x22B-v0.1 . Accessed December 15, 2024.

Irwin, V., Wang, K., Tezil, T., Zhang, J., Filbey, A., Jung, J., ... & Parker, S. (2023). Report on the Condition of Education 2023. NCES 2023-144rev. *National Center for Education Statistics*.

Jiang, A. Q., Sablayrolles, A., Roux, A., Mensch, A., Savary, B., Bamford, C., ... & Sayed, W. E. (2024). Mixtral of experts. *arXiv preprint arXiv:2401.04088*.

Jin, T., & Lu, X. (2017). A Data-Driven Approach to Text Adaptation in Teaching Material Preparation: Design, Implementation, and Teacher Professional Development. TESOL Quarterly, 52(2), 457–467. doi:10.1002/tesq.434

Kärner, T., Warwas, J., & Schumann, S. (2021). A learning analytics approach to address heterogeneity in the classroom: The teachers' diagnostic support system. Technology, Knowledge and Learning, 26(1), 31-52.

Kim, S., Shin, J., Cho, Y., Jang, J., Longpre, S., Lee, H., Yun, S., Shin, S., Kim, S., Thorne, J., & Seo, M. (2024, March 9). Prometheus: Inducing Fine-grained Evaluation Capability in Language Models. ArXiv.org. https://doi.org/10.48550/arXiv.2310.08491

Knight, K., & Marcu, D. (2002). Summarization beyond sentence extraction: A probabilistic approach to sentence compression. Artif. Intell., 139, 91-107.

Kojima, T., Gu, S. S., Reid, M., Matsuo, Y., & Iwasawa, Y. (2022). Large language models are zero-shot reasoners. *Advances in neural information processing systems*, *35*, 22199-22213

Lee, J. D., & See, K. A. (2004). Trust in automation: Designing for appropriate reliance. Human Factors, 46(1), 50–80.

Lennon, C., & Burdick, H. (2004). The lexile framework as an approach for reading measurement and success. electronic publication on www.lexile.com.

Lewkowycz, A., Andreassen, A., Dohan, D., Dyer, E., Michalewski, H., Ramasesh, V., ... & Misra, V. (2022). Solving quantitative reasoning problems with language models. Advances in Neural Information Processing Systems, 35, 3843-3857.

Li, K. C., & Wong, B. T.-M. (2021). Features and trends of personalized learning: A review of journal publications from 2001 to 2018. Interactive Learning Environments, 29(2), 182–195. https://doi.org/10.1080/10494820.2020.1811735

Li, Z., Peng, B., He, P., Galley, M., Gao, J., & Yan, X. (2023). Guiding large language models via directional stimulus prompting. *arXiv*. https://arxiv.org/abs/2302.11520

Lipovetsky, S. (2023). Readability indices structure and optimal features. Axioms, 12(5), 421. doi:10.3390/axioms12050421

Little, C. A., McCoach, D. B., & Reis, S. M. (2014). Effects of differentiated reading instruction on student achievement in middle school. Journal of Advanced Academics, 25(4), 384-402.

Masterman, T., Besen, S., Sawtell, M., & Chao, A. (2024, April 17). The Landscape of Emerging AI Agent Architectures for Reasoning, Planning, and Tool Calling: A Survey. ArXiv.org. https://doi.org/10.48550/arXiv.2404.11584

McNamara, D. S., & Graesser, A. C. (2012). Coh-Metrix: An automated tool for theoretical and applied natural language processing. In Applied natural language processing: Identification, investigation and resolution (pp. 188-205). IGI Global.

Minitab, LLC. (2024). Minitab. Retrieved from https://www.minitab.com

Mistral AI. (n.d.). *Mistral AI Large Language Models.* Retrieved from https://docs.mistral.ai/api/ . Accessed December 15, 2024.

National Center for Education Statistics. (2022). English learners in public schools. Condition of education.

Naveed, H., Khan, A. U., Qiu, S., Saqib, M., Anwar, S., Usman, M., ... & Mian, A. (2023). A comprehensive overview of large language models. *arXiv preprint arXiv:2307.06435*.

OpenAI, Achiam, J., Adler, S., Agarwal, S., Ahmad, L., Akkaya, I., … Zoph, B. (2023). GPT-4 Technical Report. doi:10.48550/ARXIV.2303.08774

OpenAI, (n.d.). *GPT-4 Turbo Model.* Retrieved from https://platform.openai.com/docs/models/gpt-4-and-gpt-4-turbo . Accessed December 15, 2024.

Papineni, K., Roukos, S., Ward, T., & Zhu, W.-J. (2002). BLEU: a Method for Automatic Evaluation of Machine Translation. https://aclanthology.org/P02-1040.pdf

Pane, J. F., Steiner, E. D., Baird, M. D., Hamilton, L. S., & Pane, J. D. (2017). Informing Progress: Insights on Personalized Learning Implementation and Effects. Research Report. RR-2042-BMGF. Rand Corporation.

Patel, N., Nagpal, P., Shah, T., Sharma, A., Malvi, S., & Lomas, D. (2023). Improving mathematics assessment readability: Do large language models help? *Journal of Computer Assisted Learning*, 39(3), 804-822. https://doi.org/10.1111/jcal.12776

Pozas, M., Letzel, V., Lindner, K. T., & Schwab, S. (2021, December). DI (differentiated instruction) does matter! The effects of DI on secondary school students' well-being, social inclusion and academic self-concept. In Frontiers in Education (Vol. 6, p. 729027). Frontiers Media SA.

Pratama, M. P., Sampelolo, R., & Lura, H. (2023). Revolutionizing education: harnessing the power of artificial intelligence for personalized learning. Klasikal: Journal of education, language teaching and science, 5(2), 350-357.

Reardon, S. F. (2018). The widening academic achievement gap between the rich and the poor. In Social stratification (pp. 536-550). Routledge.

Rello, L., Baeza-Yates, R.A., Dempere-Marco, L., & Saggion, H. (2013). Frequent Words Improve Readability and Short Words Improve Understandability for People with Dyslexia. IFIP TC13 International Conference on Human-Computer Interaction.

Sawyer, B. D., Miller, D. B., Canham, M., & Karwowski, W. (2021). Human Factors and Ergonomics in Design of A 3: Automation, Autonomy, and Artificial Intelligence. Handbook of Human Factors and Ergonomics, 1385-1416.

Shuster, K., Poff, S., Chen, M., Kiela, D., & Weston, J. (2021). Retrieval Augmentation Reduces Hallucination in Conversation. Findings of the Association for Computational Linguistics: EMNLP 2021. Presented at the Findings of the Association for Computational Linguistics: EMNLP 2021, Punta Cana, Dominican Republic. doi:10.18653/v1/2021.findings-emnlp.320

Siddharthan, A. (2014). A survey of research on text simplification. ITL – International Journal of Applied Linguistics, 165, 259-298.

Smale-Jacobse, A. E., Meijer, A., Helms-Lorenz, M., & Maulana, R. (2019). Differentiated instruction in secondary education: A systematic review of research evidence. Frontiers in psychology, 10, 472176.

Torgesen, J. K., Houston, D. D., Rissman, L. M., Decker, S. M., Roberts, G., Vaughn, S., ... & Lesaux, N. (2017). Academic literacy instruction for adolescents: A guidance document from the Center on Instruction. Center on Instruction.

Vygotsky, Lev S.(1978): Mind in Society. The Development of Higher Psychological Processes.

Watanabe, W. M., Junior, A. C., Uzêda, V. R., Fortes, R. P. de M., Pardo, T. A. S., & Aluísio, S. M. (2009). Facilita. Proceedings of the 27th ACM International Conference on Design of Communication. https://doi.org/10.1145/1621995.1622002

Wei, J., Wang, X., Schuurmans, D., Bosma, M., Chi, E., Le, Q., & Zhou, D. (2022). Chain of thought prompting elicits reasoning in large language models. arXiv. https://arxiv.org/abs/2201.11903

Wu, T., Terry, M., & Cai, C. J. (2022, March 17). AI Chains: Transparent and Controllable Human-AI Interaction by Chaining Large Language Model Prompts. ArXiv.org. https://doi.org/10.48550/arXiv.2110.01691

Wu, Y. M., Fu, J., Zhang, S., Wu, Q., Li, H., Zhu, E., Wang, Y., Yin Tat Lee, Peng, R., & Wang, C. (2023). An Empirical Study on Challenging Math Problem Solving with GPT-4. ArXiv (Cornell University). https://doi.org/10.48550/arxiv.2306.01337

Zastudil, C., Rogalska, M., Kapp, C., Vaughn, J., & MacNeil, S. (2023). Generative AI in computing education: Perspectives of students and instructors. In *arXiv [cs.HC]*. http://arxiv.org/abs/2308.04309

Zhang, T., Kishore, V., Wu, F., Weinberger, K. Q., & Artzi, Y. (2020). BERTScore: Evaluating Text Generation with BERT. ArXiv:1904.09675 [Cs]. https://arxiv.org/abs/1904.09675

Zhang, Z., Zhang, A., Li, M., & Smola, A. (2022) Automatic chain of thought prompting in large language models. arXiv preprint arXiv:2210.03493, 2022.

Zheng, L., Chiang, W.-L., Sheng, Y., Zhuang, S., Wu, Z., Zhuang, Y., Lin, Z., Li, Z., Li, D., Xing,E. P., Zhang, H., Gonzalez, J. E., and Stoica, I. (2023). Judging LLM-as-a-judge with MT-bench and Chatbot Arena. doi:10.48550/ARXIV.2306.05685

**Supplemental Materials**

1. **12th Grade Baseline Passage Samples (keywords are bolded; key phrases are italicized)**

**1.1 Black Holes (Science):**

A **black hole** is a region of spacetime where *gravity* is so powerful that nothing, including light or other electromagnetic waves, has sufficient energy to *escape*. The presence of a black hole can be inferred through its interaction with other matter and radiation, including visible light. Any matter that falls onto a black hole becomes heated by *friction*, forming **quasars**, some of the brightest objects in the universe. Stars passing too close to a supermassive black hole can be *shredded and swallowed*.

The idea of a body so big that even light could not escape was proposed by English astronomical pioneer and clergyman **John Michell** in a letter published in November 1784. Michell's simplistic calculations assumed such a body might have the same *density* as the Sun, and he concluded that one would form when a star's *diameter* exceeds the Sun's by a factor of 500 and its surface *escape velocity* exceeds the usual speed of light. Michell referred to these bodies as **dark stars**.

Scholars of the time were initially excited by the proposal that *giant but invisible* "dark stars" might be hiding in plain view. However, enthusiasm dampened when the *wavelike nature of light* became apparent in the early nineteenth century. If light were a wave rather than a particle, it was unclear what, if any, influence gravity would have on escaping light waves. In 1915, **Albert Einstein** developed his theory of **general relativity**, having earlier shown that gravity does influence light's motion.

The term "black hole" was used in print by *Life and Science News magazines* in 1963 and by science journalist **Ann Ewing** in her article "*Black Holes in Space*," dated 18 January 1964. In December 1967, a **student** reportedly suggested the phrase "black hole" at a *lecture by John Wheeler*. Wheeler adopted the term for its brevity and "advertising value," and it quickly caught on, leading some to credit Wheeler with coining the phrase.

**1.2 The Great Depression (U.S. History):**

The **Great Depression** (1929–1939) was an economic shock that impacted most countries worldwide. It was the most *prolonged and widespread* depression of the 20th century. Devastating effects were seen in both rich and poor countries, with *falling income, prices, tax revenues, and profits*. International *trade fell by more than half*, and in 1933, some *thirteen* *million Americans were unemployed*, representing 25% of the workforce and *devastating city dwellers and farmers* alike.

Unemployment had a **ripple effect** wherein the unemployed had little money to spend, and businesses lost customers and were forced to close. People lost their savings, homes, farms, businesses, and **dignity**. These *hardships* disproportionately affected American Blacks, Latinos, Indigenous people, and women, and the depression reverberated around the world. Some positive outcomes of the Great Depression can still be witnessed today: *unemployment relief, Social Security, mortgage lending practices, and the Securities Exchange Commission*.

Without income, people were *unable to pay their rent or mortgage and were evicted;* while some were lucky enough to move in with family, others *slept out in the cold on park benches, in sewers*, or in the **Hoovervilles**. These areas were full of **makeshift shanties** constructed from scrap materials, including food crates and tar paper. Metropolitan area officials often *incinerated these constructions to smoldering piles*, leaving the impoverished worse off than before. In addition to homelessness, millions faced *hunger and starvation*.

Most had previously never been dreadfully ravenous or forced to rummage through garbage, wait in bread lines, or eat at soup kitchens. Poverty-stricken rural citizens could at least grow their food, but then **natural disasters** intensified the suffering across the nation as drought, windstorms, dust storms (called black blizzards), and floods ravaged farmlands. Even farmers had no choice but to move. Subsequent depopulation impacted the local economies as communities transformed into ghost towns.

### 1.3 Vikings (World History)

**Vikings** is the modern name given to seafaring people originally from Scandinavia, which includes present-day Norway, Denmark, and Sweden. The etymology of the word "Viking" is uncertain, but during the Middle Ages, it was the designation that the Anglo-Saxons utilized synonymously for *Scandinavian* "pirates" and *raiders*. The Viking Age commenced with the earliest recorded *raids* by Norsemen *in 793* and lasted until the *Norman conquest of England in 1066*.

These Norsemen explored Europe and the edges of Asia and Asia Minor by seas and rivers for trade, raids, colonization, *and conquest*. The Vikings spoke **Old Norse** and 'wrote' inscriptions in **runes**, usually on stones or bone fragments. They had their own *laws, art, architecture*, and religion. The religion was *polytheistic*, with a belief in **many gods and goddesses**. Popular representations of the Vikings are often based on cultural clichés and stereotypes; for example, there is *no evidence that they wore horned helmets or were unclean savages*.

The Vikings traditionally survived by *farming, fishing, trapping, and hunting*. However, from the 8th to the late 11th century, Vikings sailed their **longboats** across the North Sea and Baltic Sea, down rivers across Europe, around the Mediterranean Sea, into the Black Sea, and even across the Atlantic to Iceland and North America. The territory they covered during these centuries was expansive. They *raided and pillaged,* traded, *served as mercenaries, and settled colonies* over a wide area.

Viking **settlements** have been discovered as far away as Latvia, Russia, Ukraine, and Turkey. There is archeological evidence that the Vikings traveled as far as **Baghdad**, the center of the Islamic Empire, and even as far as Uzbekistan. Two Vikings ascended to the throne of **England**—Sweyn Forkbeard (1013-1014) and his son Cnut the Great (1016-1035). **Leif Erikson**, the famed Viking who voyaged to North America, established a colony in present-day Newfoundland.

### 1.4 Simone (Biography):

**Simone Biles** is an American artistic **gymnast** born in 1997. With a total of 37 Olympic and *World Championship* **medals**, she is the *most decorated gymnast in history and is widely considered the greatest gymnast of all time*. At the 2016 Summer **Olympics** in Rio de Janeiro, Biles won individual gold medals in the all-around, vault, and floor, bronze on the balance beam, and gold as part of the United States team.

Going into the 2020 Olympic Games in Tokyo, people excitedly anticipated Biles' performance as she was expected to win at least four out of six possible gold medals. However, *toward the beginning of the competition, she withdrew,* reporting **mental health** concerns and a case of the "**twisties**," a temporary loss of air awareness while performing twisting elements. *She received extensive criticism* online and in the media, and she was even called a "quitter" who had "failed her country."

In the next couple of years, Biles became an advocate for seeking mental health treatment in sports, and she was outspoken about the impact *constant pressure and scrutiny* has on elite athletes. She returned to international competition in the 2023 World Gymnastics Championships in Antwerp, Belgium, where she **won** the individual all-around. At the U.S. Gymnastics Championships, she won the gold medal in all events and became the first gymnast to win nine all-around titles at the event.

In addition to her titles, *Biles has invented five skills* that were never performed in competition before, and these skills are named after her. At age 27, she will head to the 2024 Paris Olympics as the oldest female gymnast to compete for Team USA since 1952. Biles has *overcome many obstacles in her life and her sport*. Coaches, commentators, competitors, and fans can all agree that she is a *once-in-a-lifetime talent.*

### 1.5 The Origins of Music (Humanities):

While the origins of music remain contentious, many scholars agree that the history of music likely dates back as far as the *history of humanity* itself. Archaeologists have found **musical artifacts** such as primitive *bone flutes, rattles, and whistles* dating back to the Upper Paleolithic *40,000 years ago*. However, the origins of these instruments are likely significantly older. There is one **instrument** that has been around since the very beginning and is used in all musical traditions: **the human voice**.

Knowledge about prehistoric music is limited, but from what is known, it is typically characterized by *monophony and improvisation*. Because there was *no formal system for notating music*, melodies varied each time they were played. Simple, portable instruments were practical for hunter-gatherers. It wasn't until *humans began farming* and taking up permanent residence that new, *more complex instruments were invented,* and *systems for writing music were created*.

The cradle of **Western music** was the **Fertile Crescent**. Beginning with the development of writing, the music of literate civilizations was present in Chinese, Egyptian, Greek, Indian, Persian, Mesopotamian, and Middle Eastern societies. The earliest fragment of musical notation appears on a 4,000-year-old Sumerian **clay tablet**. It was excavated from ruins in Ugarit, **Syria**. The tablet includes **lyrics** and instrumentation instructions. Early tablets were composed in **cuneiform**, a logo-syllabic script used to write several languages of the ancient Middle East.

"Hurrian Hymn No. 6" is considered the earliest melody, but the oldest musical composition to have survived in its entirety is a first-century AD Greek song called the "Seikilos Epitaph." The emergence of the **Silk Road**, a network of Eurasian trade routes active from the second-century BCE until the mid-15th century, *aided in transmitting ideas, knowledge, and customs*. Cross-cultural exchange of musical theories, practices, and instruments was no exception.

**1.6 Disability Rights (Current Events)**

American **disability rights** have *evolved a lot over the past century*. Before the disability rights movement and before TV, **President Franklin Roosevelt** refused to be publicized in his wheelchair. He thought being a person who utilizes a *wheelchair was a weakness*. For example, while campaigning, giving speeches, or acting as a public figure, the *president hid his disability*. He perpetuated the ideology that "disability equates to **weakness.**"

This idea demonstrates and symbolizes the historic **stigma** surrounding disabilities. For a long time, disability in the United States was viewed as a personal issue, and not many political or governmental organizations existed to support individuals or their families. In the 1950s, there was a transition to volunteerism and parent-oriented organizations. One example is the **March of Dimes**. This was the start of activism and seeking support. Before this, *children with disabilities were* largely *hidden* by their *parents* out of *fear* of forced rehabilitation.

When the **civil rights** movement took off in the 1960s, disability advocates and the women's rights movements joined. Their goal was to *promote equal treatment and challenge stereotypes*. At this time, disability rights advocacy began to have a cross-disability focus. People with different kinds of disabilities (physical and mental disabilities, visual and hearing disabilities) and different essential needs came together to fight for a common cause.

It was not until 1990 that the **Americans with Disabilities Act** (ADA) was passed. This act legally *prohibited discrimination* due to disability and *mandated access* in all buildings and public areas. The ADA is historically significant in that it defined the meaning of **reasonable accommodation** to protect employees and employers. A reasonable accommodation is an *adjustment made in a system to accommodate* or make fair *the same system for an individual based on a proven need*. Needs vary from person to person.

**2. Multi-Agent Architecture Model Design:**

1. The interaction begins when the end user poses a question directed to the **Group Chat Manager** agent. This question contains not only the text to be regenerated but also the desired level of readability. The Group Chat Manager manages and coordinates with four LLM-based assistant agents: the *Selector*, the *Writer*, the *Editor* and the *Calculator*. In addition to directing the flow of communication, the Group Chat Manager is responsible for handling memory related to user interactions. This capability allows it to capture and retain valuable context regarding the user's questions and corresponding responses. The stored memory is then shared across the system, providing the other agents with context from previous interactions and ensuring more informed and relevant responses.

2. The **Selector** agent ($\mathcal{M}_p$) is called first. Its role is to reflect on the task, plan the execution plan; i.e., indicate how and how much the LLMs should be involved, and, finally, select a set of most relevant keywords and phrases to maintain the essential content of the source text. Compared to other LLM-based agents for planning, however, ours does not rely solely on the Task-Decomposition method, but instead integrates it with the processes of Refinement and Reflection. We provide this formulation in the Supplemental Materials.

- First, decompose $E$ (the environment) and $g$ (the task goal) using parameters $\Theta$ and prompt $P$ to obtain $g_i$, the task to carry out:

$$g_i = \text{decompose}(E,g;\Theta,P)$$

- Then, use $g_i$ along with $E$, $\Theta$, and $P$ to derive the sub-plan $p_i$. That is, it breaks down the task of leveling into sub-tasks, to be entrusted mainly to the agent Writer:

$$p_i = \text{sub-plan}(E,g_i;\Theta,P)$$

- The selection of the most relevant keywords ($k_i$) and phrases then takes place:

$$k_i = \text{extract\_keywords}(T;\Theta,P)$$
$$s_i = \text{rank\_sentences}(T,k_i;\Theta,P)$$

where $s_i$ is the score of sentence $i$, $T$ is the text and the extracted keywords guide the ranking process using model parameters $\Theta$ and $P$.

$$\left\{ s_{i_1}, s_{i_2}, ..., s_{i_n} \right\} = \text{top\_ranked\_sentences}(T, k_i; \Theta, P)$$

where $\left\{ s_{i_1}, s_{i_2}, ..., s_{i_n} \right\}$ are the top $n$ relevant sentences based on their scores.

- Afterwards, the Planner reflects on its sub-tasks in relation to the selected words and phrases.

$$r_i = \text{reflect}(E, g_i, p_i; \Theta, P)$$

And modify and refine the initial plan:

$$p_{i+1} = \text{refine}(E, g_i, p_i, r_i; \Theta, P)$$

3. The agent **Writer** ($\mathcal{M}_w$), therefore, writes the first version of the text, trying to meet the required level of readability, maintaining the same format and previously selected key words/phrases. The system prompt of this agent—as will be seen in the Appendix—combines the prompting technique with the few-shot of some previous examples:

$$\textbf{\textit{CoT}}\,(T_i) = \sum_{j=1}^{n} R_j(T_i)$$

where $T_i$ is the readability-modification task, and $R_j$ are the reasoning steps. This means creating a prompt that includes an example problem and the detailed chain of thought leading to the solution.

$$prompt \;=\; Example\,(T_i, \; CoT(T_i)$$

$$response_i \;=\; \mathcal{M}_w(prompt)$$

4. This is followed by the calculation of the Flesch-Kincaid Grade Level and the length of the text to ensure that the text generated by the Writer agent has met the criteria required by the user. This task is carried out by a runnable agent ($\mathcal{M}_{cal}$), which only runs the code of the FKRA_score function, a customized tool:

$$\mathcal{M}_{cal} \leftarrow fkra\_score(response_i)$$

5.  The output of the $\mathcal{M}_{cal}$ and the Writer-generated passage are then analyzed by the Editor agent ($\mathcal{M}_e$), who generates a feedback, where $p_{fb}$ is the system prompt of the Editor agent and $x$ is the chat history, including the results of the $\mathcal{M}_{cal}$ :

$$fb_t = \mathcal{M}_e(p_{fb}|\ |x|\ |response_i)$$

And checks mainly three things:

a.  Firstly, it ensures that the criteria of grade level and length of text and paragraphs have been met.
b.  It also checks whether the text has retained the most relevant keywords ($k_i$) and sentences ($S$) identified by the Planner.
c.  Finally, should the need arise, it provides some food for thought to further improve the text.

In the event that the Writer's response ($response_i$) does not meet the criteria.

Then, the response of the Editor agent would be passed to the Writer agent to refine and improve his text. This process is repeated until the task is completed:

$$response_{i+1} = \mathcal{M}_w(p_{refine}|\ |x|\ |response_i|\ |fb_t)$$

## 3. PROMPTS

Here, we provide the prompts used during the various experiments.

---

The **zero-shot** prompt we used is as follows. Divided into two sections: System Prompt and User Request. In the first part we introduce the Flesch-Kincaid Grade Level and the task to be performed. After that, we provide the target readability grade (i) $\{8^{th}||6^{th}||4^{th}\}$ and (ii) {SOURCE-TEXT} to be leveled.

| | Prompt ($P$ ) |
|---|---|
| **System** | You are a helpful writing assistant. Your task is to reduce the readability level of a source text from grade twelve to grade {$8^{th}$‖$6^{th}$‖$4^{th}$}, according to the Flesch-Kincaid Grade level.<br><br>A Flesch-Kincaid Grade Level is defined as "a measure that represents the U.S. school grade level required to understand a text," where lower scores reflect easier readability and higher scores indicate harder readability.<br><br>Ensure that the regenerated text is coherent and cohesive and maintains the division into four paragraphs of approximately 75 words each, for a total of about 300. |
| **User** | Given this text:<br><br>####{{source text}}####<br><br>Rewrite it to achieve a target Flesch-Kincaid Grade Level while keeping the original meaning and information. |

Prompts for the zero-shot solution:

| Target Grade Level | GPT-4 Turbo | Claude 3 Opus | Mixtral 8x22B |
|---|---|---|---|
| 8th-6th-4th | **SYSTEM:** You are a helpful writing assistant. Your task is to reduce the readability level of a source text from grade twelve to grade {grade}th, according to the Flesch-Kincaid Grade level.<br>　　Ensure that the regenerated text is coherent and cohesive and maintains the division into four paragraphs of approximately 75 words each, for a total of about 300.<br><br>**USER:** Q: Given this text: | You are a helpful writing assistant. Your task is to reduce the readability level of a source delimited by #### from grade twelve to grade {grade}th, according to the Flesch-Kincaid Grade level.<br><br>　　Ensure that the regenerated text is coherent and cohesive and maintains the division into four paragraphs of approximately 75 words each, for a total of about 300.<br><br>#### | You are a helpful writing assistant. Your task is to reduce the readability level of a source text after <<<>>> from grade twelve to grade {grade}th, according to the Flesch-Kincaid Grade level.<br><br>　　Ensure that the regenerated text is coherent and cohesive and maintains the division into four paragraphs of approximately 75 words each, for a total of about 300.<br><br><<< |

| | ####{text}#### Reduce the readability level to grade {grade}th, according to the Flesch-Kincaid Grade level. Ensure that the regenerated text maintains the four-paragraph division and is cohesive and coherent. Above all, it should be about 300 words long. | Source text: {inquiry}<br><br>A:<br>#### | Source text: {inquiry}<br><br>A:<br>>>> |
|---|---|---|---|

The **prompt chaining** prompts we used is as follows. The first prompt ($P_1$) aims to extract the longest sentences and keywords. It is divided into three components: the System Prompt that briefly explains the task and the target grade of readability (i) $\{8^{th}\|6^{th}\|4^{th}\}$. The few-shot section where we provide some examples of how the task should be performed, employing three different (ii) {SOURCE-TEXT #} and the corresponding (iii) {LIST-OF-SENTENCES} and (iiii) {LIST-OF-WORDS}. Then, in the final section, we offer the text to be leveled (iiiii) {SOURCE-TEXT}.

| | **Prompt ($P_1$)** |
|---|---|
| **System** | You are a helpful writing assistant. Your task is to reduce the readability level of a source text from grade twelve to grade $\{8^{th}\|6^{th}\|4^{th}\}$, according to the Flesch-Kincaid Grade level. The first task is to extract the longest sentences and words longer than 3 syllables, delimited by #### |
| **Few-shot examples** | Given this text:<br><br>####{{source text #1}}####<br>Extract the longest sentences and words (3+ syllables).<br><br>####<br>{{list of sentences #1}}<br>####<br>{{list of words #1}}<br>####<br><br>Given this text: |
| | ```
####{{source text #2}}####
Extract the longest sentences and words (3+
syllables).

####
{{list of sentences #2}}
####
{{list of words #2}}
####

Given this text:

####{{source text #3}}####
Extract the longest sentences and words (3+
syllables).

####
{{list of sentences #3}}
####
{{list of words #3}}
####
``` |
| **User** | ```
Given this text:

####{{source text}}####

Extract the longest sentences and words (3+
syllables).
``` |

After the first prompt is answered, that should provide the list of the most relevant sentences and words (i) {{Response $P_1$}}. We proceed to the second prompt ($P_2$), which is divided into three sections like the previous one: System prompt, Few-shot examples, and User request. As in the first, both the readability target grade is provided (ii) $\{8^{th}\|6^{th}\|4^{th}\}$ , as well as three pairs of examples and responses. The major difference is in the system prompt, which explains how to do the second part of the task and make use of the most relevant phrases and keywords. In the user request section, on the other hand, the answer from the previous prompt is provided.

| | **Prompt ($P_2$)** |
|---|---|
| **System** | ```
You are a helpful writing assistant. Your task is
to reduce the readability level of a source text
from grade twelve to grade {8^{th}||6^{th}||4^{th}}, according
to the Flesch-Kincaid Grade level. You will be
provided with a set containing the longest words
and sentences, delimited by ####. To perform this
``` |





| | |
|---|---|
| | task, replace the longer words with equivalent synonyms and shorten the longer sentences by splitting them into two or more sentences equivalent in length and meaning.  Ensure that the regenerated text is coherent and cohesive and maintains the division into four paragraphs of approximately 75 words each, for a total of 300. |
| **Few-shot examples** | Given this text:<br><br>####{{source text #1}}####<br><br>Reduce the readability level of the text to grade $\{8^{th}\|6^{th}\|4^{th}\}$ by shortening longer words and splitting longer sentences.<br><br>\<sentences\>:<br>####<br>{{set of sentences #1}}<br>####<br>\</sentences\><br>\<words\>:<br>####<br>{{set of words #1}}<br>####<br>\</words\><br><br>Regenerated text: {{sample #1}}<br>######<br><br>Given this text:<br><br>####{{source text #2}}####<br><br>Reduce the readability level of the text to grade $\{8^{th}\|6^{th}\|4^{th}\}$ by shortening longer words and splitting longer sentences.<br><br>\<sentences\>:<br>####<br>{{set of sentences #2}}<br>####<br>\</sentences\><br>\<words\>:<br>####<br>{{set of words #2}}<br>####<br>\</words\><br><br>Regenerated text: {{sample #2}}<br>####<br><br>Given this text: |





| | |
|---|---|
| | ```<br>####{{source text #3}}####<br><br>Reduce the readability level of the text to grade<br>{8^th‖6^th‖4^th} by shortening longer words and<br>splitting longer sentences.<br><br><sentences>:<br>####<br>{{set of sentences #3}}<br>####<br></sentences><br><words>:<br>####<br>{{set of words #3}}<br>####<br></words><br><br>Regenerated text: {{sample #3}}<br>####<br>``` |
| **User** | ```<br>Given this text:<br><br>####{{source text}}####<br><br>Reduce the readability level of the text to grade<br>{8^th‖6^th‖4^th} by shortening longer words and<br>splitting longer sentences.<br><br>####<br>{{Response P₁}}<br>####<br><br>Regenerated text:<br>``` |

The **Chain-of-Thought (CoT)** prompt we used is as follows $(P_{CoT})$.  It consists of three sections. First, we have a system prompt in which we define the task and target grade readability to be achieved (i) {$8^{th}‖6^{th}‖4^{th}$}. Then, a few-shot section in which we apply chain-of-thought prompting with three demonstrations. Each demonstration has a (ii) {SOURCE-TEXT #}, a (iii) {LIST-OF-SENTENCES} and (iiii) {LIST-OF-WORDS} and their corresponding (iiiii) {LIST-OF-SPLITTED-SENTENCES} and synonyms (iiiiii) {LIST-OF-SYNONYMS}. Finally, the User prompt with the text to be leveled (iiiiiii) {SOURCE-TEXT} .

| | **Prompt ($P_{CoT}$)** |
|---|---|
| **System** | ```<br>You are a helpful writing assistant. Your task is<br>to reduce the readability level of a source text<br>``` |





| | |
|---|---|
| | from grade twelve to grade $\{8^{th}\|6^{th}\|4^{th}\}$, according to the Flesch-Kincaid Grade level. |
| **Few-shot examples** | Q: Given this text:<br><br>####{{source text #1}}####<br><br>Reduce its readability level to $\{8^{th}\|6^{th}\|4^{th}\}$, maintaining the original division into four paragraphs of approximately 75 words and an overall length of approximately 300 words.<br><br>A: Let's think step by step. The source text has several particularly long sentences:<br>####<br>{{list of sentences #1}}<br>####<br>These could be divided into different sentences:<br>####<br>{{list of splitted sentences #1}}<br>####<br>Furthermore, we identify these words as the longest (3+ syllables):<br>####<br>{{list of words #1}}<br>####<br>Replace them with these synonyms:<br>####<br>{{list of synonyms #1}}<br>####<br><br>{{Text regenerated to level $\{8^{th}\|6^{th}\|4^{th}\}$ readability}}<br><br>Q: Given this text:<br><br>####{{source text #2}}####<br><br>Reduce its readability level to $\{8^{th}\|6^{th}\|4^{th}\}$, maintaining the original division into four paragraphs of approximately 75 words and an overall length of approximately 300 words.<br><br>A: Let's think step by step. The source text has several particularly long sentences:<br>####<br>{{list of sentences #2}}<br>####<br>These could be divided into different sentences:<br>####<br>{{list of splitted sentences #2}}<br>####<br>Furthermore, we identify these words as the longest (3+ syllables): |





```
####
{{list of words #2}}
####
Replace them with these synonyms:
####
{{list of synonyms #2}}
####

{{Text regenerated to level {8th‖6th‖4th}
readability}}

Q: Given this text:

####{{source text #3}}####

Reduce its readability level to {8th‖6th‖4th},
maintaining the original division into four
paragraphs of approximately 75 words and an overall
length of approximately 300 words.

A: Let's think step by step. The source text has
several particularly long sentences:
####
{{list of sentences #3}}
####
These could be divided into different sentences:
####
{{list of splitted sentences #3}}
####
Furthermore, we identify these words as the longest
(3+ syllables):
####
{{list of words #3}}
####
Replace them with these synonyms:
####
{{list of synonyms #3}}
####

{{Text regenerated to level {8th‖6th‖4th}
readability}}
```

| | |
|---|---|
| **User** | `Q: Given this text:`<br><br>`####{{source text}}####`<br><br>`Reduce its readability level to`<br>`{8th‖6th‖4th}, maintaining the original division into`<br>`four paragraphs of approximately 75 words and an`<br>`overall length of approximately 300 words.`<br><br>`A: Let's think step by step.` |





The **Directional Stimulus Prompting** prompt we used is as follows ($P_{DSP}$). Like the previous prompts, this one is divided into three main sections, containing a brief description of the task, examples for the few-shot section, and, finally, the user prompt in which we not only provide the {SOURCE-TEXT} to be leveled, but more importantly the {LIST-OF-SENTENCES} and {LIST-OF-WORDS} as "hints" to be offered to the model.

| | Prompt ($P_{DSP}$) |
|---|---|
| **System** | You are a helpful writing assistant. Your task is to reduce the readability level of a source text from grade twelve to grade $\{8^{th}\|6^{th}\|4^{th}\}$, according to the Flesch-Kincaid Grade level. You will be given the longest sentences and words to edit. |
| **Few-shot examples** | Q: Given this text:<br><br>####{{source text #1}}####<br><br>Reduce its readability level to $\{8^{th}\|6^{th}\|4^{th}\}$, maintaining the original division into four paragraphs of approximately 75 words and an overall length of approximately 300 words. To accomplish this task, consider splitting longer sentences and replacing longer words with credible synonyms.<br><br>Longest sentences: {{list of sentences #1}}<br>Longest words: {{list of words #1}}<br><br>A: {{Text #1 regenerated to level $\{8^{th}\|6^{th}\|4^{th}\}$ readability}}<br><br>####<br><br>Q: Given this text:<br><br>####{{source text #2}}####<br><br>Reduce its readability level to $\{8^{th}\|6^{th}\|4^{th}\}$, maintaining the original division into four paragraphs of approximately 75 words and an overall length of approximately 300 words. To accomplish this task, consider splitting longer sentences and replacing longer words with credible synonyms.<br><br>Longest sentences: {{list of sentences #2}}<br>Longest words: {{list of words #2}}<br><br>A: {{Text #2 regenerated to level $\{8^{th}\|6^{th}\|4^{th}\}$ readability}} |





| | |
|---|---|
| | ```
####

Q: Given this text:

####{{source text #3}}####
```<br><br>Reduce its readability level to $\{8^{th}\|6^{th}\|4^{th}\}$, maintaining the original division into four paragraphs of approximately 75 words and an overall length of approximately 300 words. To accomplish this task, consider splitting longer sentences and replacing longer words with credible synonyms.<br><br>```
Longest sentences: {{list of sentences #3}}
Longest words: {{list of words #3}}

A: {{Text #3 regenerated to level
``` $\{8^{th}\|6^{th}\|4^{th}\}$ ```
readability}}
``` |
| **User** | ```
Q: Given this text:

####{{source text}}####
```<br><br>Reduce its readability level to $\{8^{th}\|6^{th}\|4^{th}\}$, maintaining the original division into four paragraphs of approximately 75 words and an overall length of approximately 300 words. To accomplish this task, consider splitting longer sentences and replacing longer words with credible synonyms.<br><br>```
Longest sentences: {{list of sentences}}
Longest words: {{list of words}}

A:
``` |